\documentclass[runningheads]{llncs}
\usepackage{graphicx}

\usepackage{xcolor}
\usepackage{hyperref}

\usepackage{lipsum}

\usepackage{amsmath,amssymb,mathrsfs,amsfonts}
\usepackage{nicefrac}
\usepackage{algorithm}
\usepackage{algorithmic}

\begin{document}

\title{iPDP: On Partial Dependence Plots \\ in Dynamic Modeling Scenarios
\thanks{We gratefully acknowledge funding by the Deutsche Forschungsgemeinschaft (DFG, German Research Foundation): TRR 318/1 2021 – 438445824.}
}
\titlerunning{iPDP: On Partial Dependence Plots in Dynamic Modeling Scenarios}

\author{Maximilian Muschalik$^{*,\dagger,}$\inst{1,2}\orcidID{0000-0002-6921-0204} \and
\\Fabian Fumagalli$^\dagger$\inst{3}\orcidID{0000-0003-3955-3510} \and
\\Rohit Jagtani\inst{1}\orcidID{0009-0005-7604-256X} \and 
\\Barbara Hammer\inst{3}\orcidID{0000-0002-0935-5591} \and 
\\Eyke Hüllermeier\inst{1,2}\orcidID{0000-0002-9944-4108}}
\authorrunning{M. Muschalik et al.}
%
\institute{LMU Munich, Geschwister-Scholl-Platz 1, D-80539 München, Germany, \and
Munich Center for Machine Learning (MCML) \and
Bielefeld University, CITEC, Inspiration 1, D-33619 Bielefeld, Germany
\\
$^\dagger$ denotes equal contribution
\\
$^*$ Corresponding author: maximilian.muschalik@ifi.lmu.de}

\maketitle              

\begin{abstract}
Post-hoc explanation techniques such as the well-established partial dependence plot (PDP), which investigates feature dependencies, are used in explainable artificial intelligence (XAI) to understand black-box machine learning models. While many real-world applications require dynamic models that constantly adapt over time and react to changes in the underlying distribution, XAI, so far, has primarily considered static learning environments, where models are trained in a batch mode and remain unchanged. We thus propose a novel model-agnostic XAI framework called incremental PDP (iPDP) that extends on the PDP to extract time-dependent feature effects in non-stationary learning environments. We formally analyze iPDP and show that it approximates a time-dependent variant of the PDP that properly reacts to real and virtual concept drift. The time-sensitivity of iPDP is controlled by a single smoothing parameter, which directly corresponds to the variance and the approximation error of iPDP in a static learning environment. We illustrate the efficacy of iPDP by showcasing an example application for drift detection and conducting multiple experiments on real-world and synthetic data sets and streams.

\end{abstract}

\section{Introduction}

Since machine learning models are increasingly applied in various high-stakes environments such as healthcare \cite{Ta.2016} or energy systems \cite{GarciaMartin.2019}, models need to be explainable.
Often the best-performing models are less comprehensive than white-box alternatives resulting in a trade-off between performance and interpretability.
Explainable artificial intelligence (XAI) research addresses this trade-off by providing explanations to uncover the internal logic of such black-box models. \cite{Adadi.2018}
Model-agnostic XAI methods can be applied to any machine learning model regardless of its structure.
The model is treated as a black-box and is systematically probed with differently structured inputs to observe its output behavior.
A special kind of global, model-agnostic explanation is the visualization of the \emph{feature effects}. 
Feature effect methods such as the \emph{Partial Dependence Plot} (PDP) \cite{Friedman.2001} aim at marginalizing a model's output along a feature axis.
This allows for visually inspecting the possibly complicated relationship between a model's outputs and feature values.

XAI research has traditionally focused on static learning environments, where models are trained on fixed data sets and ought to be stationary.
However, machine learning models are increasingly applied in different dynamic learning environments such as \emph{incremental learning} \cite{Losing.2018} from data streams or \emph{continual lifelong learning} \cite{Parisi.2019}.
For instance, predictive maintenance models are often fitted on a constant stream of sensor information \cite{Davari.2021} or financial services providers benefit from online credit scoring models \cite{Jillian2020}.
Such application scenarios often require incremental models to be updated efficiently one by one sequentially.
Online models learn continuously from an ever-evolving stream of information where the learning task or the environment may change over time.
Shifts in the data distributions, called \emph{concept drift}, may arise from failing sensors or irregular readings in predictive maintenance applications \cite{Davari.2021}, or because of pandemic-induced lockdowns in energy forecasting systems \cite{Rouleau.2021} or hospital admission criteria \cite{Duckworth.2021}.

Similar to the static batch learning setting, high-stakes online learning applications require XAI approaches to enable the use of high-performance black-box models.
However, in dynamic environments with ever-changing models, more than static explanations are required.
To properly explain dynamic models at any point in time, special XAI approaches are required that are, like the incremental models, updated over time one by one \cite{Fumagalli.2022,Muschalik.2022}. 
In this work, we are interested in the well-established PDP to explain the dependencies of features in the model.
The PDP is able to uncover changes in the model, which may remain undiscovered by measures based on changes in accuracy \cite{Muschalik.2022} or global feature importance \cite{Fumagalli.2022,muschalik2023isage}.
We, thus, provide an incremental variant of PDP, referred to as iPDP, that efficiently computes a stream of PDPs over time.
This stream can be viewed as an \emph{explanation stream} or \emph{interpretability stream}, which summarizes the data and the model efficiently at any point in time.
In our experiments, we construct a synthetic data stream to highlight the benefits of iPDP over simple feature importance based methods.
We further provide an example application that shows how this explanation stream of iPDP can be used to reliably detect changes in a real-world data stream setting.

\paragraph{Contribution.}
Our main contributions include:
\begin{itemize}
    \item We introduce iPDP as a novel, model-agnostic explanation method that naturally retrieves the feature effects in non-stationary modeling scenarios, such as online learning from data streams.
    \item We establish important theoretical guarantees for iPDP, such as that iPDP reacts properly to real drift and, in a static environment, approximates the PDP.
    \item We demonstrate and validate the efficacy of iPDP by conducting experiments on synthetic and real-world online learning scenarios.
    \item We implement iPDP as part of an online learning XAI Python package\footnote{iPDP is part of the \emph{iXAI} framework at \url{https://github.com/mmschlk/iXAI}.}.
\end{itemize}

\paragraph{Related Work.}

With the increasing use of streaming data, there is a growing need to develop methods that can accurately explain dynamic models. 
Since tree-based approaches are commonly applied in such streaming scenarios model-specific approaches that compute global feature importance (FI) have been proposed \cite{Cassidy.2014,Gomes.2019}.
Moreover, also model-agnostic variants exist that calculate FI for any model type trained on real-time streams of data \cite{Fumagalli.2022,muschalik2023isage}.
Yet, the approach of using a single point estimate to model feature importance can conceal the underlying effects of a feature across the entire feature space, which could reveal additional insights.
This could further enrich approaches aiming at describing concept drift through means of explanation \cite{Haug.2022,Hinder.2023,Muschalik.2022}.

For static, non-streaming, environments, several techniques to visualize single-feature effects exist.
The PDP shows the marginal effect a set of features has on the predicted outcome of a machine learning model \cite{Friedman.2001}. 
A PDP can show whether the relationship between the target and a feature is linear, monotonic or more complex. \cite{Friedman.2001,Herbinger.2022,molnar2022}
PDPs have successfully been applied in various application domains across disciplines \cite{berk2013statistical,Elith2008,moosbauer2022explaining,Zhao2022}.
Application domains include, imbalanced classification costs as encountered in the criminal justice domain \cite{berk2013statistical}, animal-habitat factors in ecological research \cite{Elith2008}, congestion prediction for traffic planning \cite{Zhao2022}, and hyperparameter optimization for automated machine learning pipelines \cite{moosbauer2022explaining}.

Apart from visualizing the feature effects, the PDP can also be summarized in a FI score by computing the deviation between the individual feature scores in relation to the mean PD curve \cite{Greenwell.2018,molnar2022}.
When significant interaction effects are present, the relationship between the response and predictors may vary considerably.
Consequently, the use of an average curve, such as the PDP, may obscure the intricate nature of the modeled relationship, thereby masking the underlying complexity \cite{Herbinger.2022}. 
Individual Conditional Expectation (ICE) plots \cite{Goldstein.2015} refine the PDP by graphing the functional relationship between the predicted response and the feature for individual observations.
Accumulated Local Effects (ALE) curves \cite{Apley2020} present an alternative visualization approach to PDPs, which do not require unreliable extrapolation with correlated predictors.
ALE plots are far less computationally expensive than PDPs.
However, in case of strong correlation, an interpretation of the effect across intervals is not permissible \cite{molnar2022}.
To overcome the problem of feature interactions, stratifying PDPs by conditioning on a correlated and potentially interacting feature to group ICE curves was suggested \cite{Grömping2020}.
VINE \cite{Britton.2019} achieves this by clustering ICE curves with similar slopes.
REPID \cite{Herbinger.2022} uses a tree-based approach to identify and cluster homogeneous ICE curves to produce individual effect curves for each unique cluster. 
Shapley dependence plots \cite{lundberg2019consistent} use the SHAP value of a feature for the y-axis and the value of the feature for the x-axis. 
Thus, SHAP dependence plots capture vertical dispersion due to interaction effects in the model.

\section{Theoretical Background}
In the following, we introduce the notion of explaining black-box models through the features' effects as constructed with the PDP (Section~\ref{section_feature_effects}) and briefly establish the problem setting of fitting time-dependent models on data streams (Section~\ref{section_online_learning}).

\paragraph{Notation.} Given a $d$-dimensional feature space $\mathcal{X} \in \mathbb{R}^d$ and a target space $\mathcal{Y}$ (e.g., $\mathcal{Y} = [0,1]^c$ for a $c$-dimensional classification problem or $\mathcal{Y} = \mathbb{R}$ in case of regression) the corresponding machine learning model aims to learn a prediction function $f: \mathcal{X} \rightarrow \mathcal{Y}$.
We denote $(X^1,\dots,X^d)$ and $Y$ as the corresponding random variables for the feature and target spaces contained in the joint data distribution $\mathbb{P}(X,Y)$. 
A dataset $\mathcal{D} = \{(\mathbf{x}_i,y_i)\}_{i=1}^{n}$ consists of $n$ samples drawn i.i.d. from $\mathbb{P}(X,Y)$.
We denote the the $i$-th observation as $\mathbf{x}_i = \left(x^1_i,\dots,x^d_i\right)$ and the realizations of the $j$-th feature $X^j$ as $\mathbf{x}^j = \left(x^j_1,\dots,x^j_n\right)$.
We further denote a set of features indices with $S \subseteq \{1,\dots,d\}$ and its complement as $\bar{S} = S^\complement$ and abbreviate the corresponding random variables with $X^S$ and $X^{\bar{S}}$.

\subsection{Estimating Feature Effects with Partial Dependence}
\label{section_feature_effects}

\begin{figure}[t]
    \centering
    \includegraphics[height=0.25\textheight]{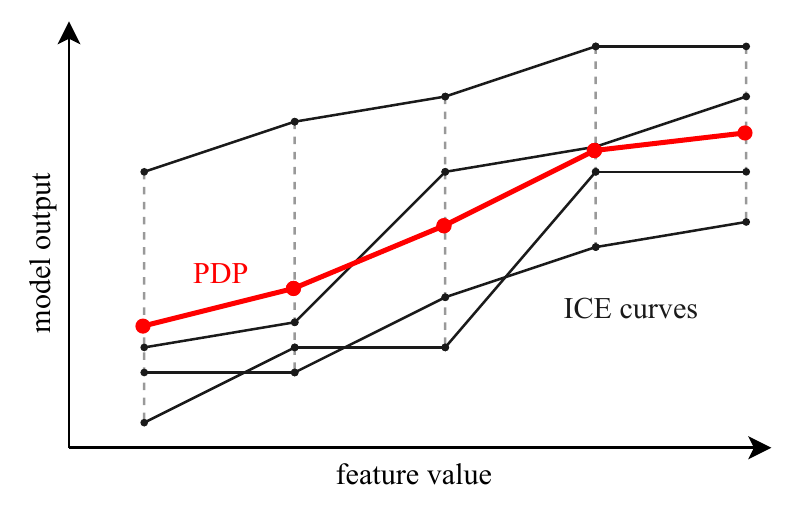}
    \caption{Illustration of a PDP that reveals a positive effect for an arbitrary feature and model. The black lines denote individual ICE curves and the red line the corresponding PDP.}
    \label{fig_PDP_Illustration}
\end{figure}

Different approaches exist for retrieving the relationship of a feature on the underlying model's predictions.
The \emph{Partial Dependence Plot} (PD Plot, or PDP) \cite{Friedman.2001} is a well-established method for retrieving the marginal effect of features on a model. \cite{Herbinger.2022,molnar2022,Molnar.2021b}
The PD function of a feature set $S$ is defined as
\begin{equation*}
    f_S^{\text{PD}}(\mathbf{x}^S) = \mathbb{E}_{X^{\bar{S}}}\left[f(\mathbf{x}^S,X^{\bar{S}})\right].
\end{equation*}
The PD function marginalizes the underlying model over all features in $\bar{S}$.
Since $\mathbb{P}$ is unknown, the PD function can be approximated via Monte-Carlo integration using all observed data points 
\begin{equation}\label{equation_pdp_estimate}
    \hat{f}_S^{\text{PD}}(\mathbf{x}^S) = \frac{1}{n} \sum_{i=1}^{n} f(\mathbf{x}^S,\mathbf{x}^{\bar{S}}_i).
\end{equation}
The approximation therefore evaluates the model on the set of all $n$ observations, where the values for features in $\bar{S}$ stem from the original data points and the feature values in $S$ are replaced by values in $\mathbf{x}^S$.
To visualize the PDP, $m$ grid points $\mathbf{x}^S_1,\dots,\mathbf{x}^S_m$ are used to construct a PD curve using the pairs $\{\mathbf{x}^S_k,\hat{f}_S^{\text{PD}}(\mathbf{x}_k^S)\}_{k=1}^m$.
Grid points can be created in equidistance on the feature scale or based on the feature distribution \cite{Molnar.2021}. 
As an additional layer of interpretation, the PDP can be enriched with ICE curves \cite{Goldstein.2015}.
ICE curves show each observation's trajectory across the feature space revealing more complicated relationships because of feature interactions or correlations \cite{Goldstein.2015,Herbinger.2022,Molnar.2021b}, i.e. a single term in the sum of (\ref{equation_pdp_estimate}).
The PDP and its ICE curves are illustrated in Fig.~\ref{fig_PDP_Illustration}.

Since, values for features in $S$ are sampled independently of $\mathbf{x}^{\bar{S}}$, synthetic data points can be created that break the dependence between features in $S$ and $\bar{S}$.
Hence, the model is evaluated with unrealistic data points, which is also referred to as being \emph{off-manifold} \cite{Frye.2021} or sampled by the \emph{marginal} expectation \cite{Janzing.2020}.

\subsection{Online Learning from Data Streams under Drift}
\label{section_online_learning}

In many online learning settings from data streams, a dynamic model $f_t$ learns from observations arriving sequentially over time such that the stream at time $t$ consists of observations $(x_0,y_0), (x_1,y_1) \dots (x_t,y_t)$.
The model $f_t$ is incrementally updated with each new observation $(x_t,y_t)$ resulting in a new model $f_t \rightarrow f_{t+1}$. \cite{Gama.2014,Losing.2018,Lu.2018}
Compared to traditional batch learning settings, where models are trained on an accessible and static dataset, learning from data streams entails various challenges.
First, data streams yield unbounded sets of training data resulting in new observations arriving in the future. 
Second, time intervals between new data points may be short, such that incremental models need to be updated efficiently to cope with the high frequency of arriving observations.
Traditionally, the high capacity and frequency of the data prohibits exhaustively storing the complete stream.
Hence, the model is ought to be evaluated and fitted only once on each new observation.
Incremental updates can be realized by conducting a single gradient update for neural networks or linear/logistic regression models \cite{Losing.2018}, or by updating split nodes in incremental decision trees \cite{Bifet.2009,Domingos.2000,Hulten.2001}.
Lastly, in most streaming scenarios, the data generating process is considered to be non-stationary leading to so-called \emph{concept drift}. \cite{Gama.2014,Lu.2018}

\paragraph{Concept Drift.} In general, concept drift can be defined as a shift in the joint distribution of the data generating process, i.e. there exist two time points $t_1,t_2$ such that $\mathbb{P}_{t_1}(X,Y) \neq \mathbb{P}_{t_2}(X,Y)$ \cite{Gama.2014}.
Applying the Bayes rule to the joint distribution, concept drift can be further decomposed into $\mathbb{P}(X,Y) \propto \mathbb{P}(X) \mathbb{P}(Y\mid X)$.
A change in the feature's distribution $\mathbb{P}(X)$ without affecting the dependencies between $X$ and $Y$ is referred to as \emph{virtual drift}.
Virtual drift thus, theoretically, does not affect the decision boundaries of a trained model $f$.
In contrast, \emph{real drift} refers to a change in the conditional distribution $\mathbb{P}(Y \mid X)$, which necessitates to adapt the learned model to reflect the novel functional dependency. \cite{Lu.2018}

A shift can occur because of a real change in the functional relationship between the targets and features like exogenous events such as pandemic-induced lockdowns on energy consumption patterns \cite{Rouleau.2021} or hospital admission criteria \cite{Duckworth.2021}.
Data distributions may shift smoothly from one concept into the other (gradual drift), or transition rapidly (sudden drift) \cite{Lu.2018}.
The \emph{effect} of a particular feature for predicting the target values may change substantially in all variants of concept drift.

\section{Incremental Partial Dependence Plots}
In an incremental learning setting on a data stream the model is updated with every observation and may change fundamentally over time, if concept drift occurs.
Providing insights into the model $f_t$ at every time step with measures such as the PDP is a challenging, yet important, task to understand how the model's reasoning changes over time.
In Section \ref{section_PDP_shifts}, we discuss the effect of concept drift on PDPs and identify two important challenges, caused by real and virtual concept drift.
We then present, in Section \ref{section_iPDP}, a novel and efficient algorithmic approach to compute incremental PDPs over time using minimal computational resources.
Our approach results in an \emph{interpretability} or \emph{explainability} stream, which provides a stream of PDP values alongside the data stream, which can be further used in applications to understand how feature dependencies change over time.
In Section \ref{section_iPDP_theory}, we analyze our approach theoretically and provide meaningful guarantees, which further support our algorithmic approach.

\begin{figure}[t]
    \hspace{1pt}
    \includegraphics[width=0.99\textwidth]{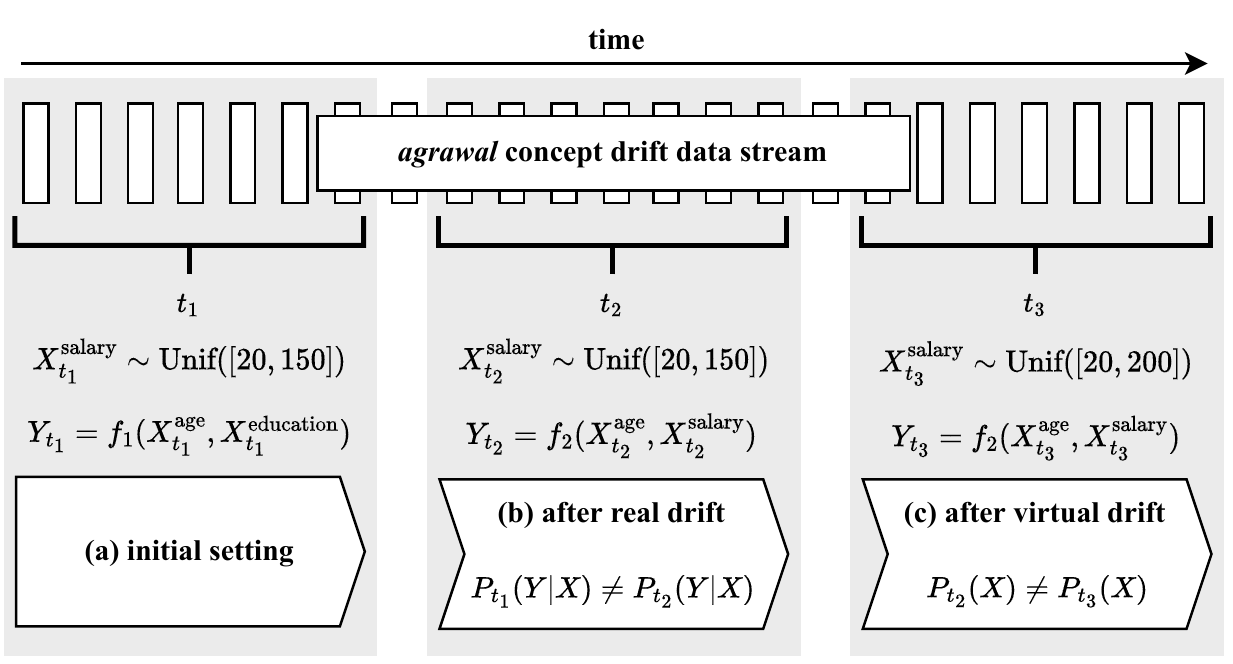}
    \\[-1em]
    \begin{minipage}[c]{\textwidth}
    \hspace{1.5pt}
    \includegraphics[width=0.30\textwidth]{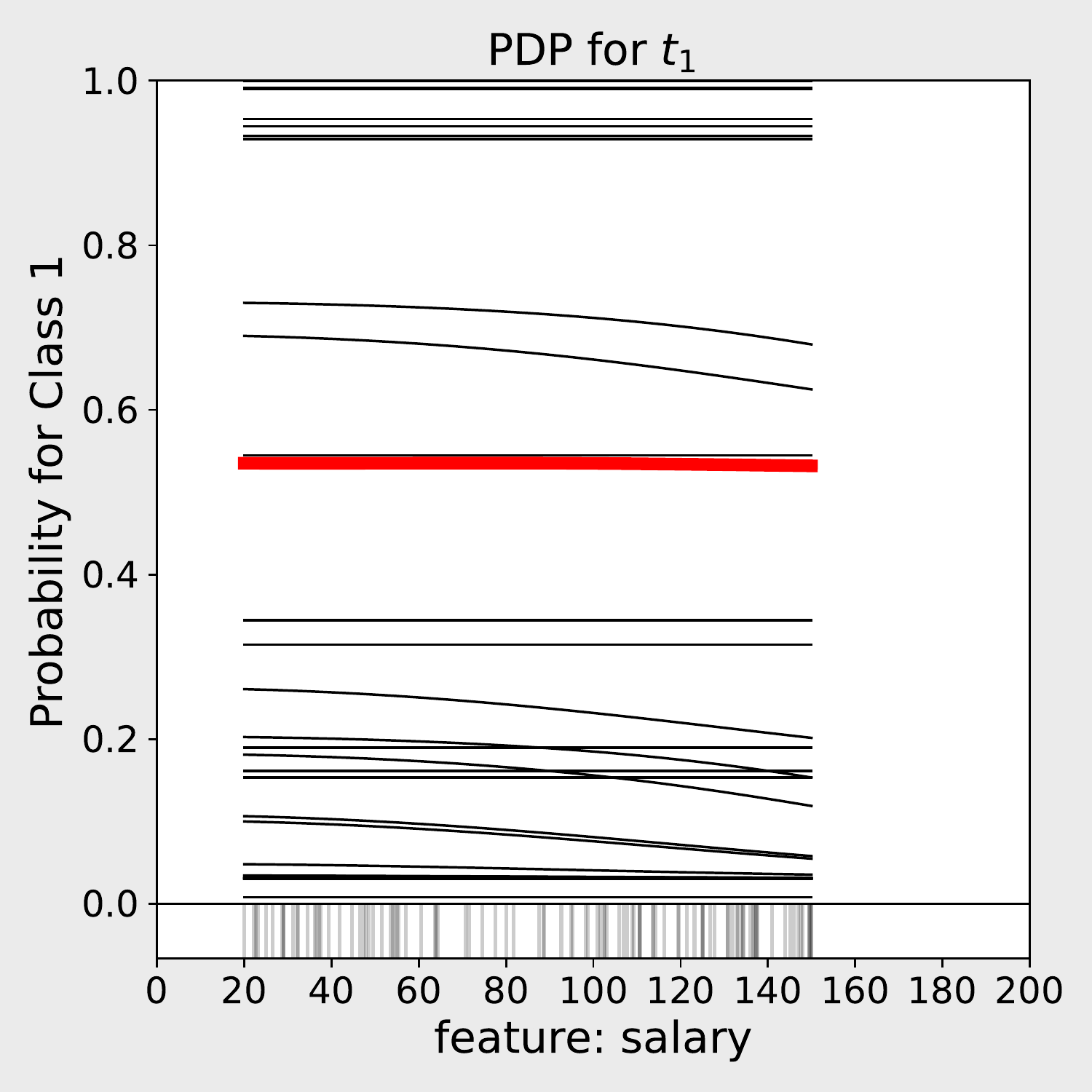}
    \hspace{7.5pt}
    \includegraphics[width=0.30\textwidth]{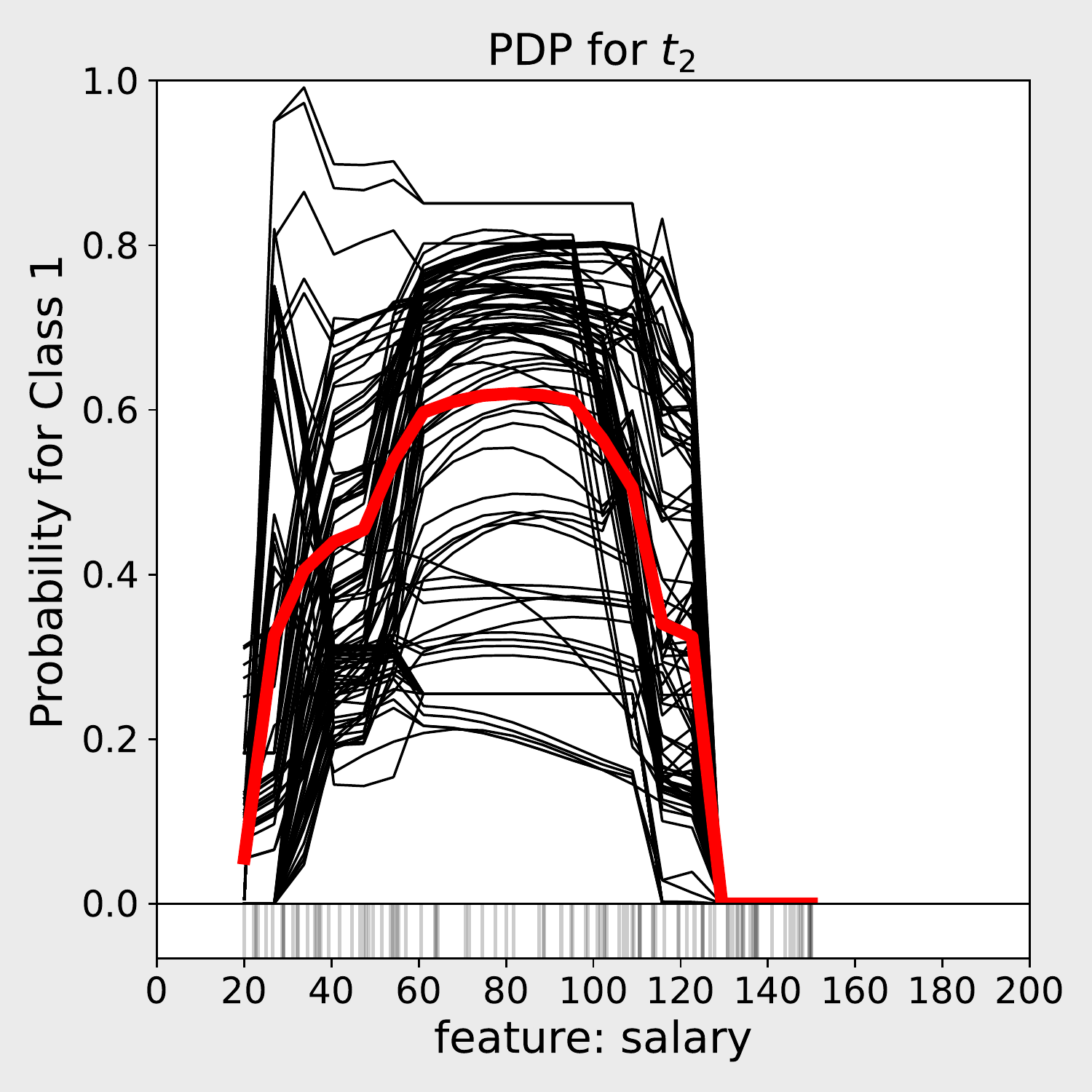}
    \hspace{8pt}
    \includegraphics[width=0.30\textwidth]{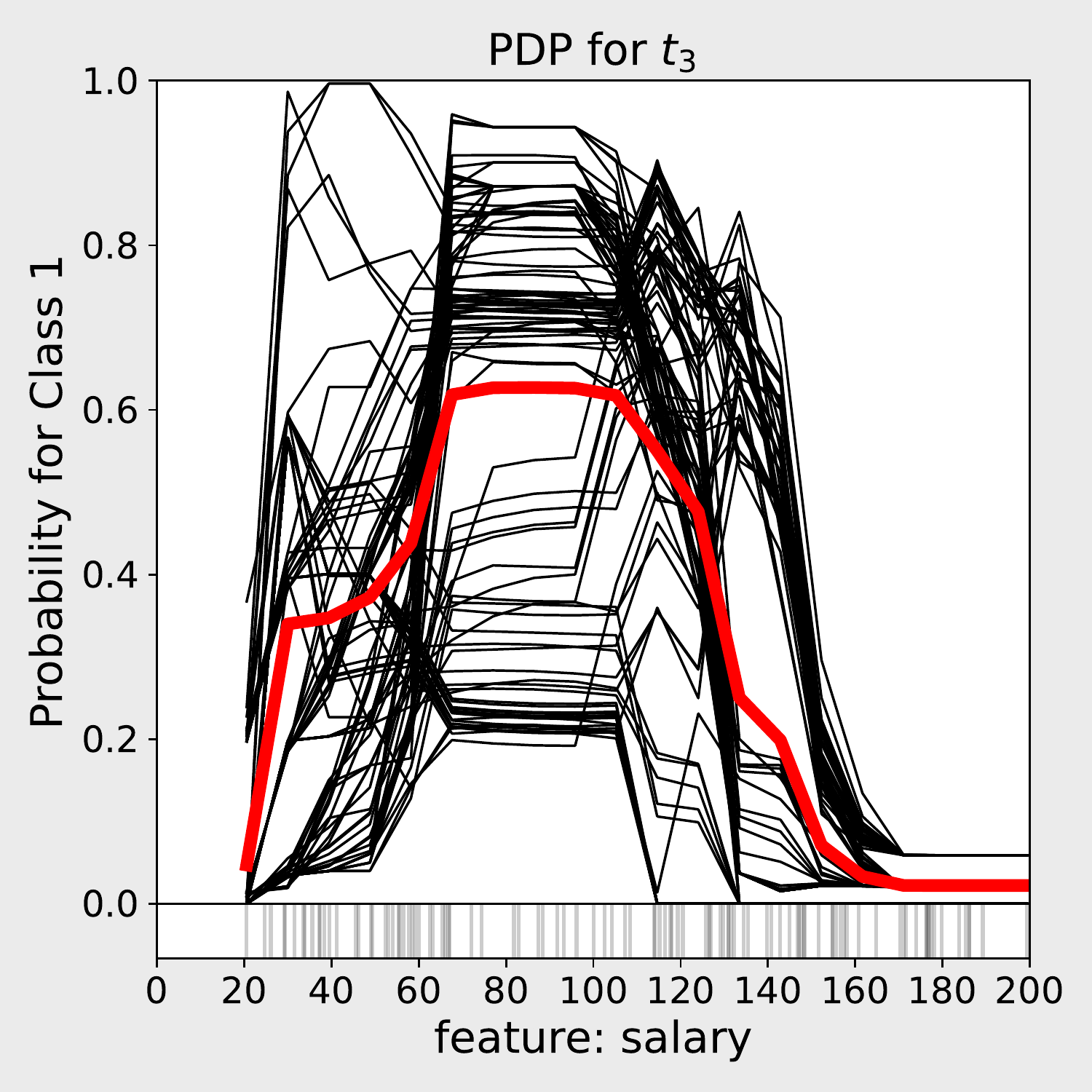}
    \end{minipage}
    \caption{An incremental model is fitted on a synthetic data stream based on the \emph{agrawal} \protect{\cite{Agrawal.1993}} concepts. The stream contains two concept drifts between three time intervals ($t_1$, $t_2$, $t_3$). The PDPs computed for the incremental model illustrate the drifts. At first the feature \emph{salary} is not relevant for the classification task (a). Then a \emph{real drift} (b) changes the classification problem and the feature becomes relevant. Lastly, a \emph{virtual drift} (c) leads to a new feature range. Here, $f_1(X_t^{\text{age}},X_{t}^{\text{education}})$ and $f_2(X_t^{\text{age}},X_t^{\text{salary}})$ refer to the \emph{agrawal} classification functions and not a model.}
    \label{fig_pdp_on_stream}
\end{figure}

\subsection{Partial Dependence Plots under Concept Drift}\label{section_PDP_shifts}

We consider the incremental learning scenario described in Section \ref{section_online_learning}, where we observe data points over time and update the current model with every new observation.
Assuming that the model sufficiently approximates the underlying data generating distribution, we can distinguish between virtual and real drift and its effect on the PDPs over time.

\paragraph{PDP under real drift.}
Real drift is usually reflected in a change in the model's decision boundaries.
Such a change will often yield a change in the shape of the PDP, when compared with previously calculated PDPs, as illustrated in Fig.~\ref{fig_pdp_on_stream}.
Clearly, under real drift the model may change substantially and previously calculated PDPs will provide misleading insights into the model's behavior. 
A time-dependent PDP measure should therefore, ideally, recompute the PDP after every model update.
However, in practice when working with rapid data streams, repeated computations quickly become infeasible, as at every time step computing (\ref{equation_pdp_estimate}) requires $n$ model evaluations.
To improve computational efficiency, previous model evaluations may be used but recent evaluations should be favored over outdated ones \cite{Fumagalli.2022,Muschalik.2022}.

\paragraph{PDP under virtual drift.}
In the static batch setting, the PDP is commonly calculated and shown based on equidistant grid points within the feature's range.
While observing new data points, the feature's range may change due to virtual drift, as illustrated in Fig.~\ref{fig_pdp_on_stream}.
Virtual drift will, ideally, not change the decision boundary of a learned model, but instead change the grid points in which the PDP is visualized.
A time-dependent PDP measure should therefore maintain grid points in the \emph{current} range of the features and \emph{forget} about outdated feature ranges.

\subsection{Incremental Partial Dependence Plot}

\label{section_iPDP}
\begin{algorithm}[t]
\caption{iPDP Explanation Procedure}
\label{alg_iPDP}
\begin{algorithmic}[1]
\REQUIRE stream $\left\{\mathbf{x}_{t},y_{t}\right\}_{t=1}^{\infty}$, model $f_t(.)$, feature set of interest $S$, smoothing parameter $0 < \alpha \leq 1$, number of grid points $m$, and storage object $R_t$
\STATE initialize $\hat{f}_S^{\text{PD}}(\mathbf{x}^S_{0,k},1) \gets 0$
\FORALL{$(\mathbf{x}_t, y_t) \in$ stream}
    \STATE $\{\tilde{\mathbf{x}}_{t,k}^S \}_{k=1}^m \gets \textsc{GetGridPoints}(R_t, m)$ \COMMENT{e.g., equidistant points, quantiles, etc.}
    \FOR{$k = 1, \dots, m$} 
        \STATE $\mathbf{x}^S_{t,k} \gets (1-\alpha) \cdot \mathbf{x}^S_{t-1,k} + \alpha \cdot \tilde{\mathbf{x}}^S_{t,k} $ \COMMENT{update grid point}
        \STATE $\hat{y}_k \gets f_t\left(\tilde{\mathbf{x}}^S_{t,k},\mathbf{x}^{\bar{S}}_{t}\right)$ \COMMENT{evaluate on model evaluation point}
        \STATE $\hat{f}_S^{\text{PD}}(\mathbf{x}^S_{t,k},t) \gets (1-\alpha) \cdot \hat{f}_S^{\text{PD}}(\mathbf{x}^S_{t-1,k},t-1) + \alpha \cdot \hat{y}_k$ \COMMENT{update point-wise estimates}
    \ENDFOR
    \STATE $R_t \gets \textsc{UpdateStorage}(R_{t-1}, x_t^S)$ \COMMENT{add $x_t^S$ to the storage object}
    \STATE \textbf{Output:} $\frac{\hat{f}_S^{\text{PD}}(\mathbf{x}^S_{t,k},t)}{1 - (1 - \alpha)^t}, \frac{\mathbf{x}^S_{t,k} }{1 - (1 - \alpha)^t}$ \COMMENT{debiasing of estimates and grid points} 
\ENDFOR
\end{algorithmic}
\end{algorithm}

\noindent We now present an efficient algorithmic approach to maintain time-dependent PDPs, i.e. we construct a PD curve at every time step $t$ based on the pairs $\{\mathbf{x}^S_{t,k},\hat{f}_S^{\text{PD}}(\mathbf{x}^S_{t,k},t)\}_{k=1}^m$, where we denote $\mathbf{x}^S_{t,k}$ as the grid points constructed at time $t$.
The main iPDP procedure is described in Algorithm~\ref{alg_iPDP}, which includes a \emph{debiasing} factor\footnote{The debiasing factor ensures that for a constant sequence the exponential moving average remains constant and will be theoretically justified in Section~\ref{section_iPDP_theory}.} in the output $(1-(1-\alpha)^t)^{-1}$ for both $\mathbf{x}^S_{t,k}$ and $\hat{f}_S^{\text{PD}}(\mathbf{x}^S_{t,k},t)\}$.
We further distinguish between the \emph{final grid points} $\mathbf{x}^S_{t,k}$ used for visualization of iPDP and \emph{temporary model evaluation points} $\tilde{\mathbf{x}}^S_{t,k}$ used to evaluate the model at time $t$.
Our approach is based on two mechanisms:  At every time step, we compute $k=1,\dots,m$ point-wise estimates for $\hat f_S^{\text{PD}}(\mathbf{x}^S_{t,k},t)$ by updating the previous estimates $\hat f_S^{\text{PD}}(\mathbf{x}^S_{t-1,k},t-1)$ using the current model evaluated at the model evaluation point $f_t(\tilde{\mathbf{x}}^S_{t,k},\mathbf{x}_t^{\bar S})$, accounting for real drift.
The grid points $\mathbf{x}^S_{t,1},\dots\mathbf{x}^S_{t,m}$ used for visualization are then constructed by updating the previous grid points $\mathbf{x}^S_{t-1,1},\dots\mathbf{x}^S_{t-1,m}$ based on the model evaluation point $\tilde{\mathbf{x}}^S_{t,k}$ obtained from the current feature's range, accounting for virtual drift.
We now first describe the updating mechanism for the point-estimates and then discuss our approach of updating the grid points dynamically.

\begin{figure}[t]
    \centering
    \includegraphics[height=0.25\textheight]{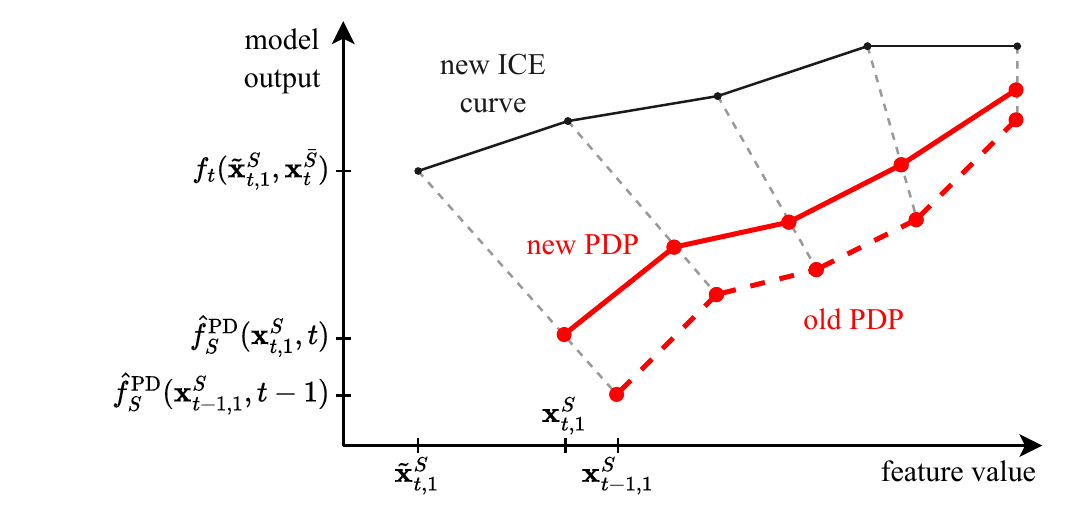}
    \caption{The current PDP (red, dashed) is updated with a new ICE curve (black), which spans over a new feature range resulting in the updated PDP (red, solid) to slowly transition into the new feature range.}
    \label{fig:iPDP_Illustration}
\end{figure}

\paragraph{Updating point-wise iPDP estimates $\hat{f}_S^{\text{PD}}(\mathbf{x}^S_{t,k},t)$.}
Clearly, recomputing (\ref{equation_pdp_estimate}) at every time step using the current model $f_t$ quickly becomes infeasible, as it requires $n$ model evaluations for every grid point and feature set.
Instead, to re-utilize previous model evaluations, we compute an exponential moving average for each grid point $\mathbf{x}^S_{t,k}$ (further described below), using the recursion
\begin{equation}\label{equation_iPDP_grid}
    \hat{f}_S^{\text{PD}}(\mathbf{x}^S_{t,k},t) := (1-\alpha) \cdot \hat{f}_S^{\text{PD}}(\mathbf{x}^S_{t-1,k},t-1) + \alpha \cdot f_t(\tilde{\mathbf{x}}^S_{t,k},\mathbf{x}^{\bar{S}}_{t})
\end{equation}
with $\hat{f}_S^{\text{PD}}(\mathbf{x}^S_{0,k},0) := 0$ for $k=1,\dots,m$ and smoothing parameter $0<\alpha<1$.
Here, $\mathbf{x}_{t,k}^S$ refers to the $k$-th grid point at time $t$, $\tilde{\mathbf{x}}^S_{t,k}$ refers to the $k$-th model evaluation point in the current feature's range (further described below), whereas $\mathbf{x}_t$ refers to the observed data point at time $t$.
The model evaluations $\{f_t(\tilde{\mathbf{x}}^S_{t,k},\mathbf{x}^{\bar{S}}_{t})\}_{k=1}^m$ can be viewed as an ICE curve of the current observation $\mathbf{x}_t$, and hence iPDP uses an exponential moving average of ICE curves of previous observations, as illustrated in Fig.~\ref{fig:iPDP_Illustration}.
The main difference to (\ref{equation_pdp_estimate}) is that instead of evaluating $f_t$ for $n$ data points, we evaluate $f_t$ for \emph{only one} data point and utilize the previous calculations, which greatly reduces computational complexity.
In Section \ref{section_iPDP_theory}, we theoretically justify our smoothing approach over grid points and model evaluation points by relating it to a local linearity assumption of $f_t$ in the range of model evaluation points $\{\tilde{\mathbf{x}}^S_{s,k}\}_{s=1}^t$. 

\paragraph{Updating iPDP grid points $\mathbf{x}^S_{t,k}$.}
Similarly to smoothing the point-wise estimates, we average each previous grid point $\mathbf{x}^S_{t-1,k}$ with a \emph{model evaluation point} $\tilde{\mathbf{x}}^S_{t,k}$ (further described below) based on the current feature's distribution using an exponential moving average.
Formally, the final iPDP grid points used for visualization are
\begin{equation}\label{equation_grid_points_recursion}
    \textbf{iPDP grid point: }\mathbf{x}^S_{t,k} := (1-\alpha) \cdot \mathbf{x}^S_{t-1,k}+ \alpha \cdot  \tilde{\mathbf{x}}^S_{t,k}
\end{equation}
at each time step $t$ for $k=1,\dots,m$ and initial grid point $\mathbf{x}^S_{0,k} := 0$.

\paragraph{Dynamically maintaining model evaluation points $\tilde{\mathbf{x}}^S_{t,k}$.}
In the static batch setting, it is common practice to compute the PDP on equidistant grid points within a  \emph{minimum} and \emph{maximum} value, which is obtained from the feature's distribution.
These values can be obtained based on the actual minimum and maximum of a feature's range or based on their \emph{quantile} distribution, e.g. $5\%$ and $95\%$ quantile \cite{Herbinger.2022,Molnar.2021}.
In a dynamic learning environment with virtual drift, the feature's range may change significantly over time.
We therefore maintain the minimum $\mathbf{x}^S_{t,\min}$ and maximum $\mathbf{x}^S_{t,\max}$ of the model evaluation points based on the feature's distribution dynamically over a recent time frame.
Formally, the equidistant model evaluation points within $[\mathbf{x}^S_{t,\min},\mathbf{x}^S_{t,\max}]$ are then given as
\begin{equation*}
   \textbf{Model evaluation point: } \tilde{\mathbf{x}}^S_{t,k} := \mathbf{x}^S_{t,\min}+ \frac {k-1}{m-1}( \mathbf{x}^S_{t,\max}- \mathbf{x}^S_{t,\min}),
\end{equation*}
at each time step $t$ for $k=1,\dots,m$.
Our efficient implementation (Algorithm~\ref{alg_extrem_value_storage}) of the rolling maximum (and minimum) is based on the observation that previous data points with lower values than a more recent observation can be discarded.
We further provide an efficient implementation to store frequency distributions of features.
Traditional streaming histograms \cite{BenHaim.2010} summarize the whole stream's distribution and cannot be applied to store rolling distributions, since old observations outside the window's range need to be forgotten.
Instead, our frequency storage (Algorithm~\ref{alg_geom_reservoir_storage_efficient}) extends on the geometric reservoir sampling procedure \cite{Fumagalli.2022} by allowing observations to be skipped (i.e., $0<p_{\text{inc}} <1$) and enforcing an ordering on the reservoir.
Based on this frequency storage, quantiles can be dynamically computed to obtain the $\mathbf{x}^S_{t,\min}$ and $\mathbf{x}^S_{t,\max}$ values.
Furthermore, it is possible to use the dynamic quantiles instead as grid points depending on the application scenario \cite{Herbinger.2022,Molnar.2021}.

\subsection{Theoretical Guarantees}\label{section_iPDP_theory}

In this section, we provide theoretical results that further support our algorithmic approach.
In particular, we show that iPDP, in expectation, is a weighted sum of previous PD function evaluations, which uses higher weight on recent evaluations to properly react to real drift, as required in Section \ref{section_PDP_shifts}.
We further motivate the debiasing factor $(1-(1-\alpha)^t)^{-1}$ in Algorithm~\ref{alg_iPDP}, which corrects the bias in a static environment.
All proofs can be found in Appendix \ref{appendix_proofs}.

\paragraph{Approximation of a time-dependent PD function.}
In a dynamic environment, such as a data stream with concept drift, the data-generating distribution $(X_t,Y_t)$ and the model $f_t$ are not static and possibly noisy.
Our first result shows that iPDP computes, in expectation, a weighted average of PD function evaluations, where recent evaluations receive a higher weight.

\begin{theorem}\label{theorem_expectation}
    iPDP reacts to real drift and favors recent PD values, as 
    \begin{equation*}
    \mathbb{E}[\hat f_S^{\text{PD}}(\mathbf{x}^S_{t,k},t)] 
    = \alpha \sum_{s=1}^{t} (1-\alpha)^{t-s}\underbrace{\mathbb{E}_{X_s^{\bar S}}\left[f_s(\tilde{\mathbf{x}}_{s,k}^S,X^{\bar{S}}_s)\right]}_{\text{PD function at time }s}, \text{ for } k=1,\dots,m.
\end{equation*}
\end{theorem}

By Theorem \ref{theorem_expectation}, iPDP estimates an exponential average of PD function evaluations, where recent evaluations receive a higher weight, which allows iPDP to react to real drift, as required in Section \ref{section_PDP_shifts}.

\paragraph{Guarantees in static environments.}

To establish further theoretical results, we restrict our analysis to a \emph{static environment}, i.e. observations $(x_t,y_t)$ are iid from a static distribution $\mathbb{P}(X,Y)$ and the model remains fixed over time $f \equiv f_t$.
Clearly, these assumptions imply that the PD function does not change over time.
If the sequence of $\{\tilde{\mathbf{x}}^S_{s,k}\}_{s=1}^t$ remains constant, then we can show that the expectation of $\hat f_S^{\text{PD}}(\mathbf{x}^S_{t,k},t)$ is biased, which motivates the debiasing factor $(1-(1-\alpha)^t)^{-1}$.
\begin{theorem}\label{theorem_static_fixed_expectation}
Let observations $(x_0,y_0),\dots,(x_t,y_t)$ be iid from $\mathbb{P}(X,Y)$ and $f \equiv f_t$ be a static model. Let further the sequence of model evaluation points be static, i.e. $\tilde{\mathbf{x}}_{k} \equiv \tilde{\mathbf{x}}_{s,k}$ for $s=1,\dots,t$. Then,
\begin{equation*}
    \mathbb{E}[\hat f_S^{\text{PD}}(\mathbf{x}^S_{t,k},t)] = \left(1-(1-\alpha)^t \right) f_S^{\text{PD}}(\tilde{\mathbf{x}}^S_{k}),
\end{equation*}
which motivates the debiasing factor for $\hat f_S^{\text{PD}}(\mathbf{x}^S_{t,k},t)$ in Algorithm~\ref{alg_iPDP}.
\end{theorem}
Clearly, assuming a fixed sequence of model evaluation points is restrictive. 
When no virtual drift is present, we may instead assume that the collection of temporary model evaluation points $\{\tilde{\mathbf{x}}^S_{s,k}\}_{s=1}^t$ remains in a close range.
If these points remain close, then it is reasonable to assume that $f$ behaves \emph{locally linear} within this range.
We can then show that the expectation of iPDP at time $t$ is actually the PD function evaluated at the final grid point $\mathbf{x}^S_{t,k}$ used for visualization.

\begin{theorem}\label{theorem_static_expectation}
  Let observations $(x_0,y_0),\dots,(x_t,y_t)$ be iid from $\mathbb{P}(X,Y)$ and $f \equiv f_t$ be a static model.
  If $f$ is locally linear in the range of temporary model evaluation points $\{\tilde{\mathbf{x}}^S_{s,k}\}_{s=1}^t$ for $k=1,\dots,m$, then 
  \begin{equation*}
      \mathbb{E}\left[\hat f_S^{\text{PD}}(\mathbf{x}^S_{t,k},t)\right] = f_S^{\text{PD}}\left(\mathbf{x}^S_{t,k}\right) \text{ and } \mathbb{E}\left[\frac{\hat f_S^{\text{PD}}(\mathbf{x}^S_{t,k},t)}{1-(1-\alpha)^t}\right] =  f_S^{\text{PD}}\left(\frac{\mathbf{x}^S_{t,k}}{1-(1-\alpha)^t}\right),
  \end{equation*}
  where the second equation justifies the debiasing factor for $\mathbf{x}^S_{t,k}$ in Algorithm~\ref{alg_iPDP}.
\end{theorem} 
We can further prove that the variance of iPDP is directly controlled by the smoothing parameter $\alpha$ and, in case of local linearity, controls the approximation error.
 
\begin{theorem}\label{theorem_static_variance}
Let observations $(x_0,y_0),\dots,(x_t,y_t)$ be iid from $\mathbb{P}(X,Y)$ and $f \equiv f_t$ be a static model. Then, the variance is controlled by $\alpha$, i.e. $\mathbb{V}[\hat f_S^{\text{PD}}(\mathbf{x}^S_{t,k},t)] = \mathcal O(\alpha)$.
In particular, if $f$ is locally linear in the range of temporary model evaluation points $\{\tilde{\mathbf{x}}^S_{s,k}\}_{s=1}^t$ for $k=1,\dots,m$, then for every $\epsilon >0$
\begin{align*}
    \mathbb{P}\left(\vert \hat f_S^{\text{PD}}(\mathbf{x}^S_{t,k},t)-f_S^{\text{PD}}(\mathbf{x}^S_{t,k}) \vert > \epsilon\right) = \mathcal O(\alpha).
\end{align*}
\end{theorem}

\section{Experiments and Applications}
\label{section_experiments}

We evaluate and show example use cases of iPDP in different experimental scenarios.
First in section~\ref{section_experiment_change_detection}, we explore how iPDP can be used to create a data stream of explanations, which can be further refined and used for dynamic explanations or drift detection.
Second, in section~\ref{section_experiment_dynamic_environments}, we demonstrate how iPDP can be used in a dynamic learning environments to recover up-to-date feature effects which would be obfuscated by online FI measures.
In section~\ref{section_experiment_static}, we validate our theoretical result and show how iPDP converges to the ground truth batch PDP in a static learning environment.\footnote{All experiments are based on \emph{sklearn}, \emph{pytorch} and the \emph{river} online learning framework. All datasets are publicly available and described in the supplement \ref{sec_appendix_data_set_description}. The code to reproduce the experiments can be found at
\url{https://github.com/mmschlk/iPDP-On-Partial-Dependence-Plots-in-Dynamic-Modeling-Scenarios}.}

\subsection{Use Case: Change Explanation based on iPDP}
\label{section_experiment_change_detection}

\begin{figure}[t]
    \centering
    \begin{minipage}[c]{0.32\textwidth}
        \includegraphics[width=\textwidth]{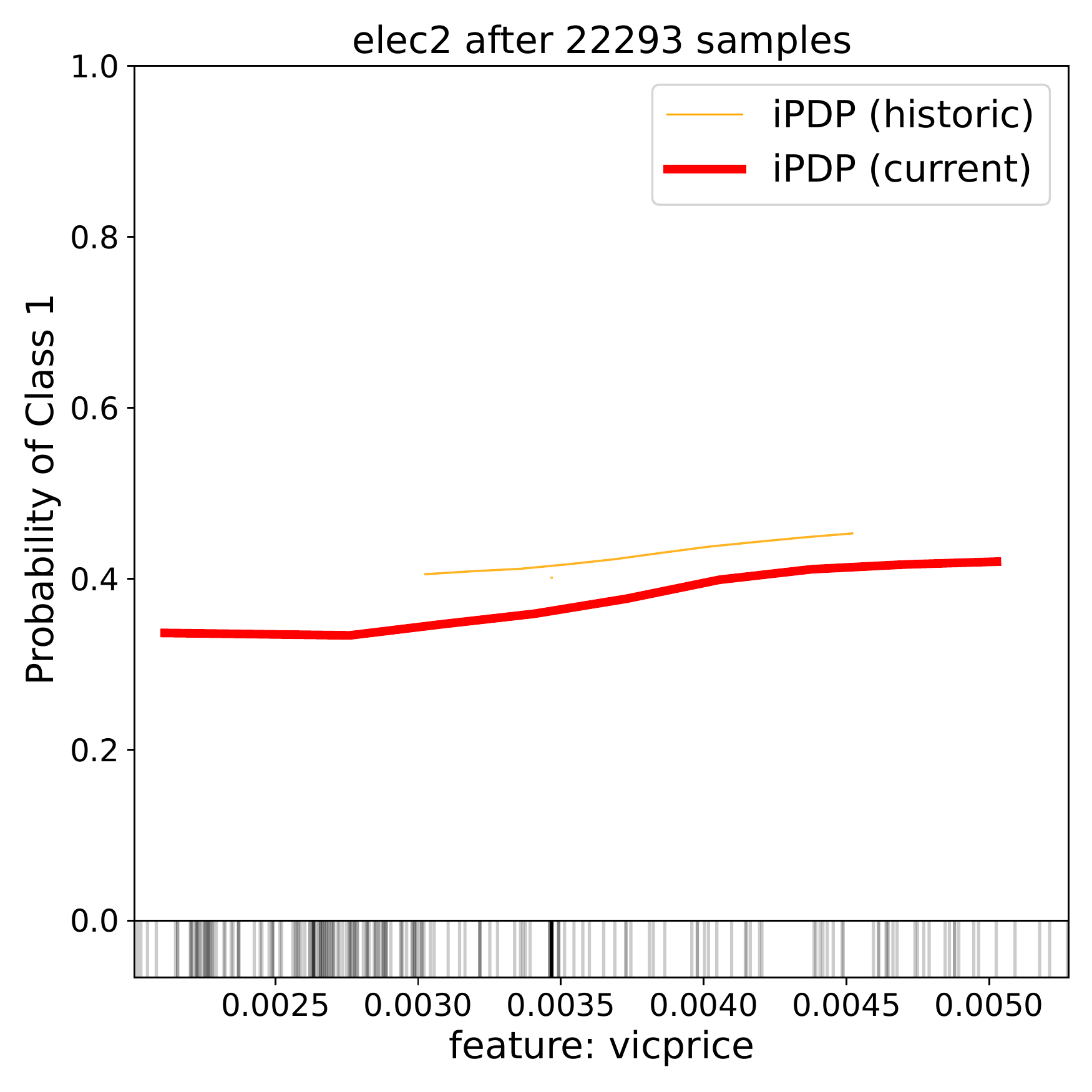}
    \end{minipage}
    \hfill
    \begin{minipage}[c]{0.32\textwidth}
        \includegraphics[width=\textwidth]{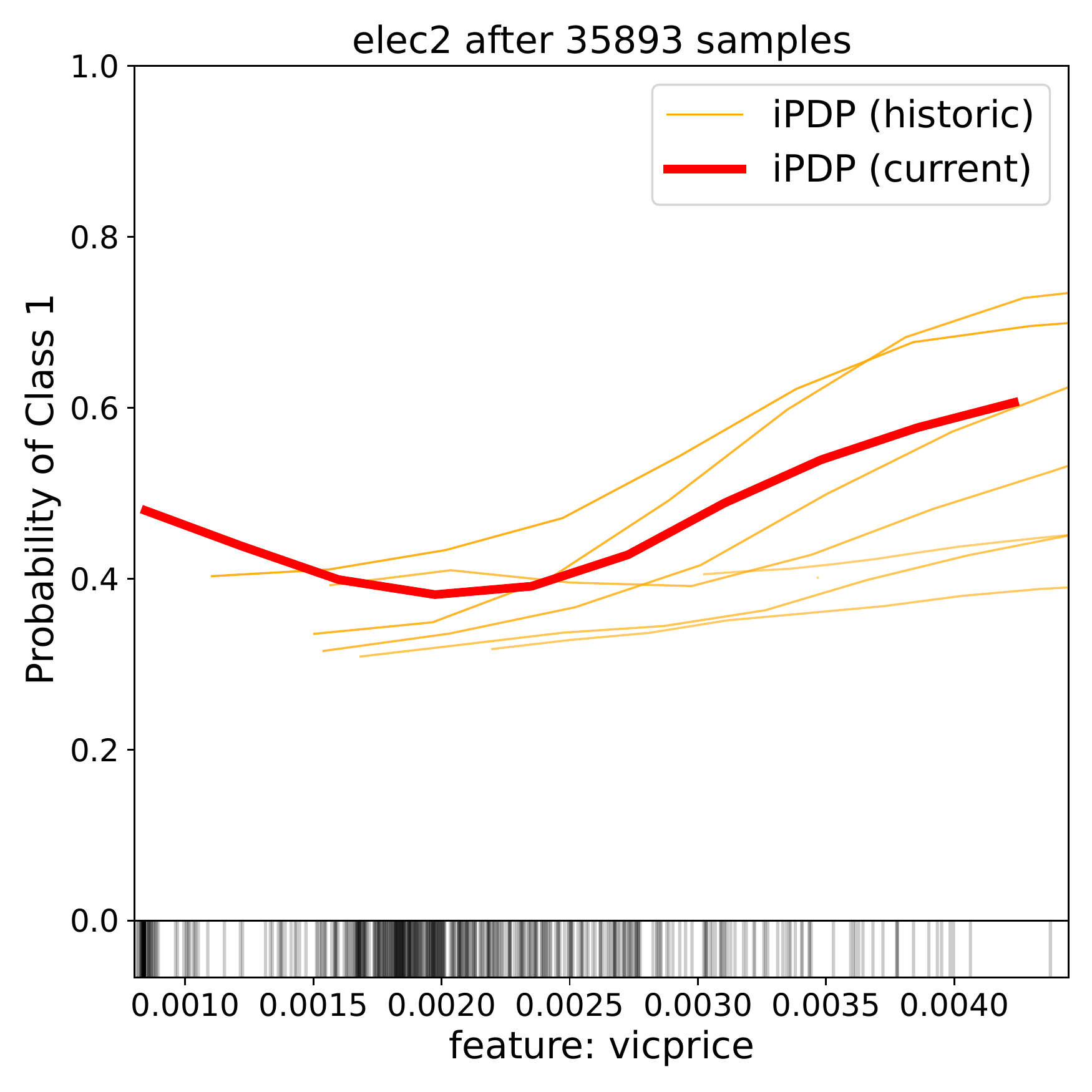}
    \end{minipage}
    \hfill
    \begin{minipage}[c]{0.32\textwidth}
        \includegraphics[width=\textwidth]{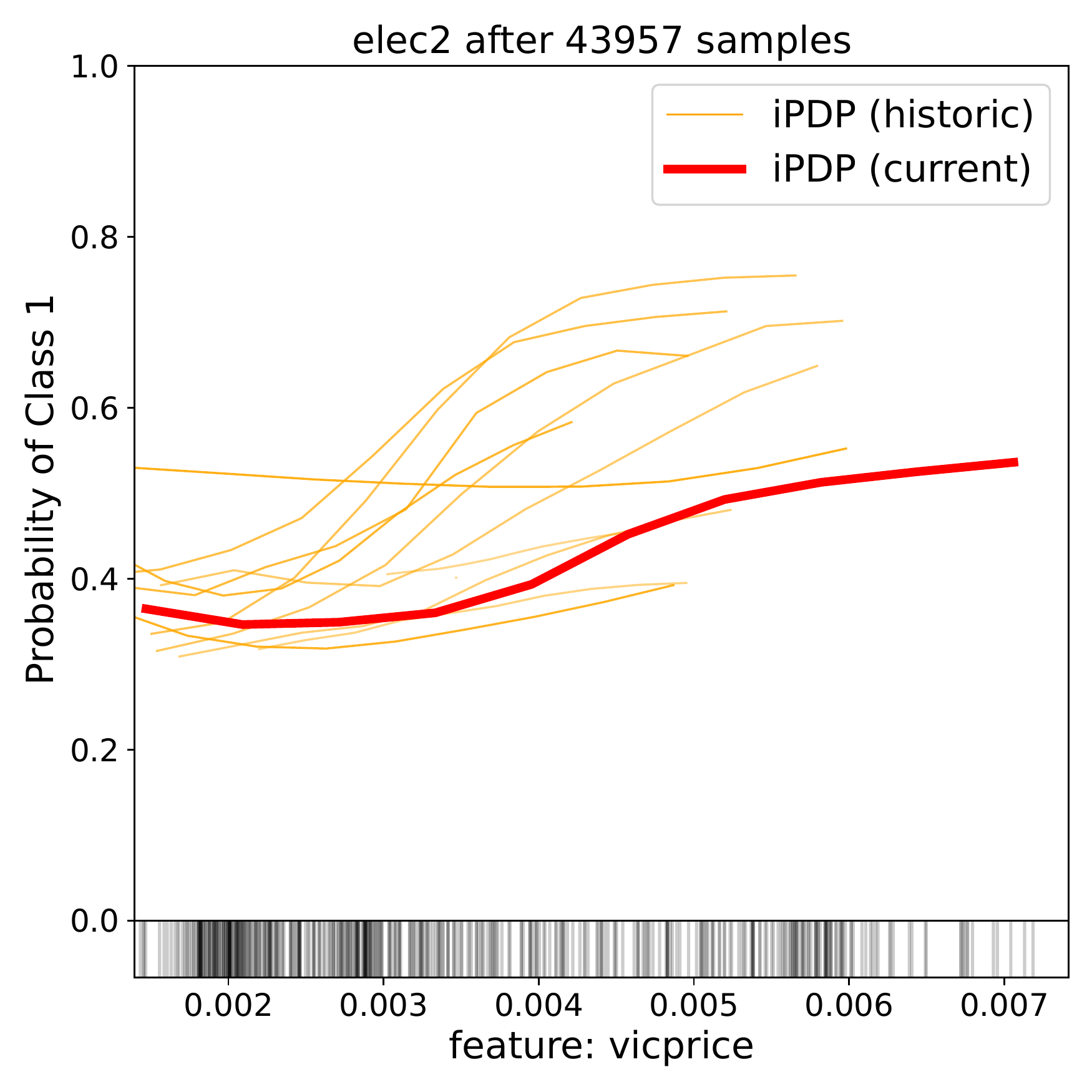}
    \end{minipage}
    \includegraphics[width=\textwidth]{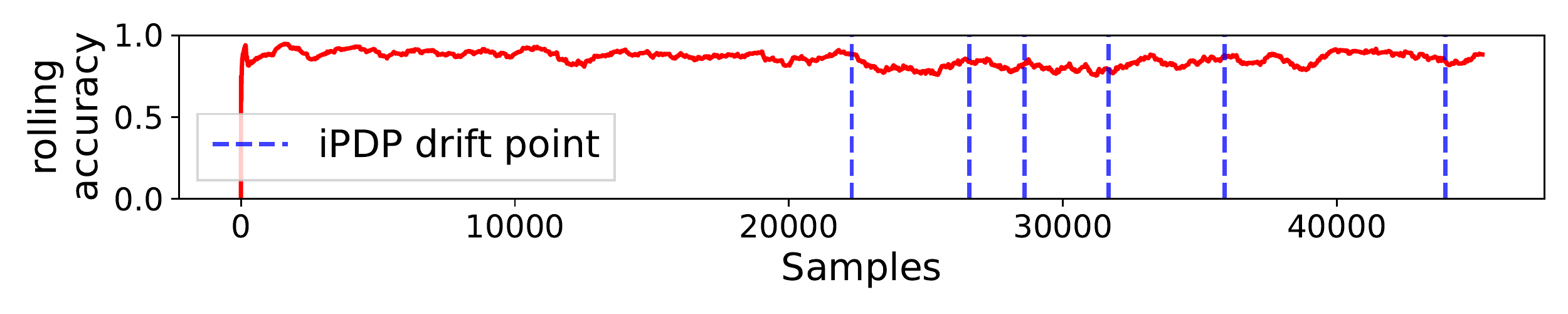}
    \caption{Three iPDP explanations at different time points (after $22\,293$, $35\,893$, and $43\,957$ observations) for an ARF fitted on the \emph{elec2} data stream. The iPDPs are computed for the \emph{vicprice} feature. The time points for generating the explanations are detected by ADWIN based on the iPDP (blue lines).}
    \label{fig_change_detection_explanations}
\end{figure}

iPDP creates time-dependent feature effect curves at any time.
Therein, iPDP transforms the data stream on which a model is incrementally fitted into a stream of explanations.
This stream of dynamic effect curves can be analyzed and monitored similar to other streams of information.
Traditionally, concept drift detection algorithms are applied on the stream of model performance values \cite{Lu.2018}.
Yet, some performance-based drift detectors like ADWIN \cite{Bifet.2007} can be applied on any numerical values streamed over time.
Hence, we propose to analyze the explanation stream of incremental effect curves with classical drift detectors.
This methodology can be used to detect and investigate changes in the underlying model more directly than by relying on the performance metrics \cite{Muschalik.2022}.
\\
To illustrate, how the iPDP stream can be monitored continuously, we fit and explain an Adaptive Random Forest (ARF) \cite{Gomes.2017} with 10 trees as base learners on the well-established \emph{elec2} \cite{elec2.1999} concept drift stream.
We compute the iPDP with $\alpha = 0.001$ over $10$ grid points yielding a curve at any time for each feature.
The grid points are equidistant between quantiles $Q_{5}$ and $Q_{95}$ as derived by Algorithm~\ref{alg_geom_reservoir_storage_efficient}.
To summarize the $10$ individual grid points of the time-dependent PD curve, we condense the iPDP into a single FI score.
The PDP-based importance score can be calculated as the deviation of each individual feature value from the average PD curve \cite{Greenwell.2018,molnar2022}.
We apply ADWIN as a change detector on this stream of FI scores.
With every drift detected by ADWIN (blue lines in Fig.~\ref{fig_change_detection_explanations}), we plot the current iPDP and present the current model behavior to the user.
Fig.~\ref{fig_change_detection_explanations} shows a selection of three iPDP explanations at three different time points identified by ADWIN as change points.
It shows, how the model behaves differently at different points in time.
Until approximately $22\,000$ samples, the model has no effect given the \emph{vicprice} feature of the stream, since the feature's value was constantly zero.
After the concept drift, the \emph{vicprice} value become relevant for the model in terms of a U-shaped effect on the predictions.
Lastly, for a larger feature range the effect translates into a more linear relationship.
A monitoring system like this can be used for various application domains, where decisions must be automated based on certain model characteristics.

\subsection{iPDP in Dynamic Learning Environments}
\label{section_experiment_dynamic_environments}

\begin{figure}[t]
    \centering
    \begin{minipage}[c]{0.32\textwidth}
        \includegraphics[width=\textwidth]{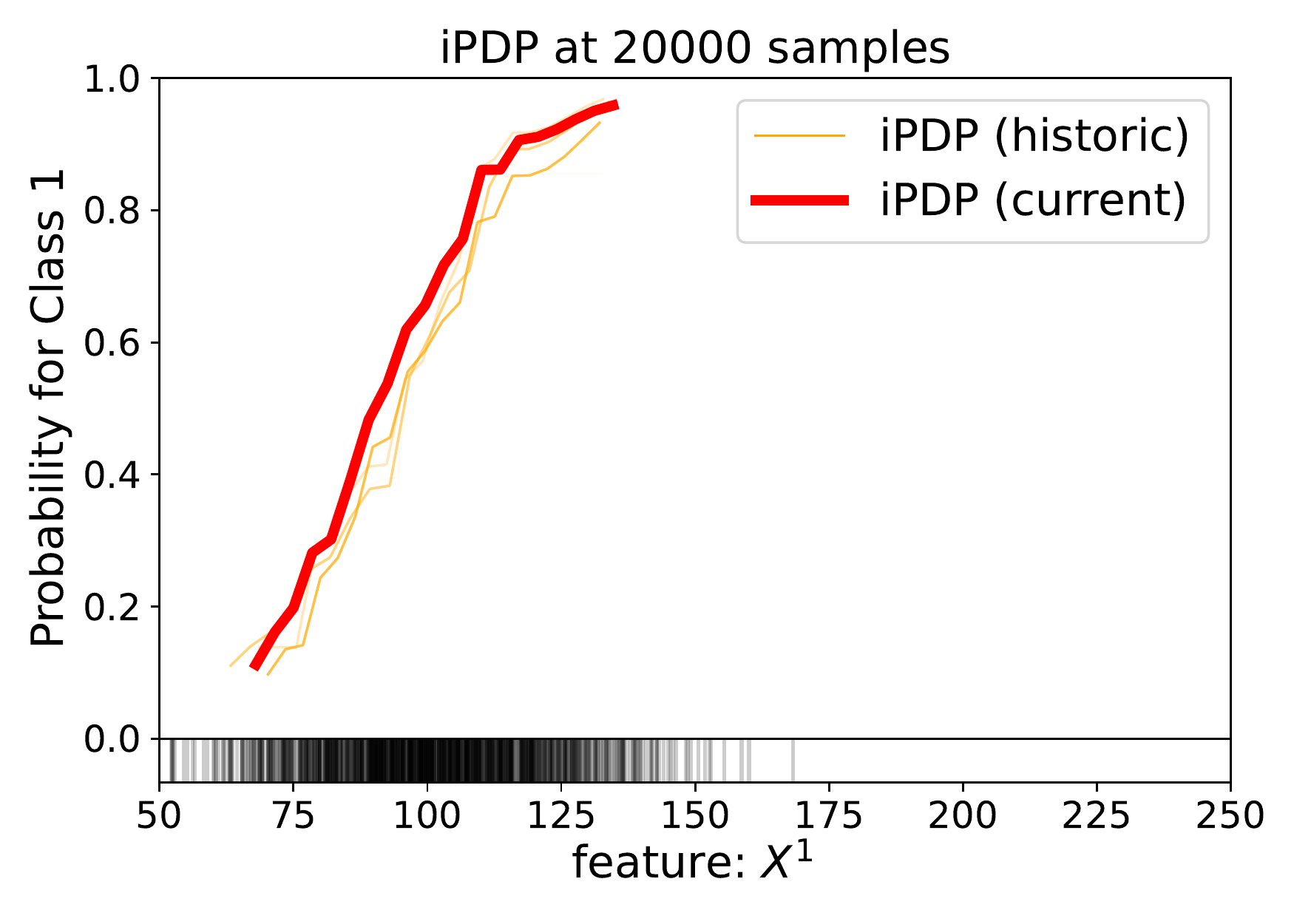}
    \end{minipage}
    \hfill
    \begin{minipage}[c]{0.32\textwidth}
        \includegraphics[width=\textwidth]{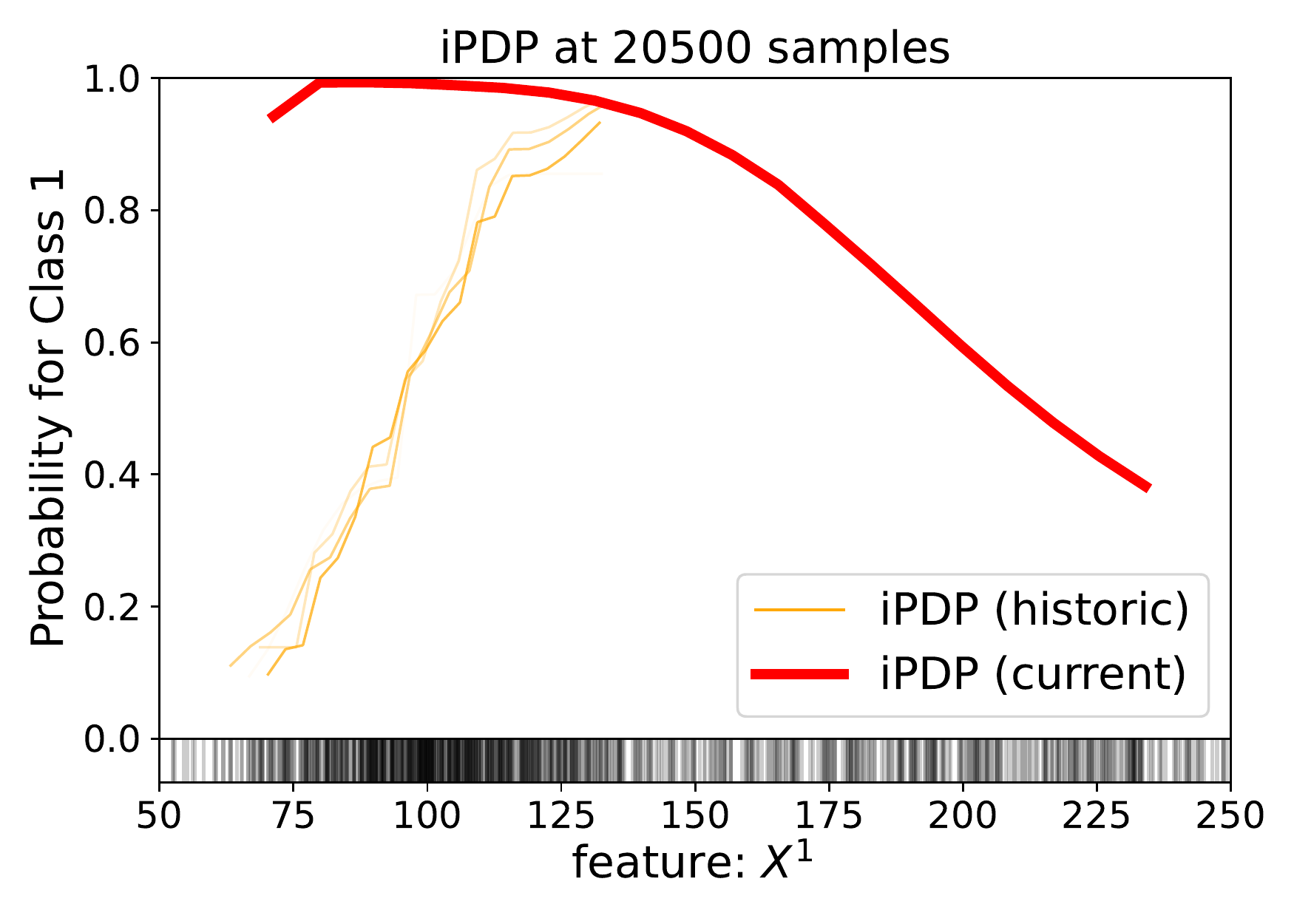}
    \end{minipage}
    \hfill
    \begin{minipage}[c]{0.32\textwidth}
        \includegraphics[width=\textwidth]{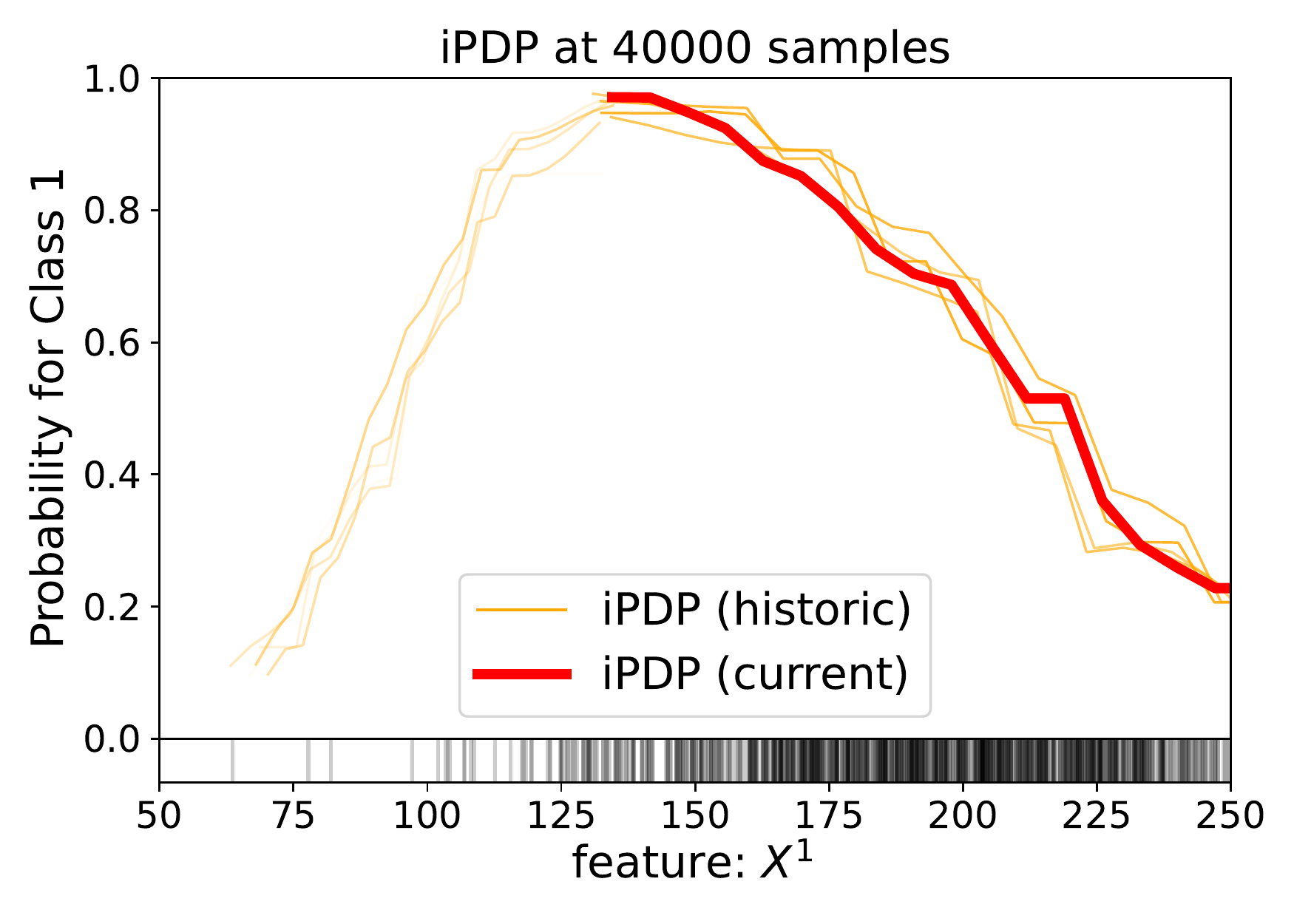}
    \end{minipage}
    \includegraphics[width=\textwidth]{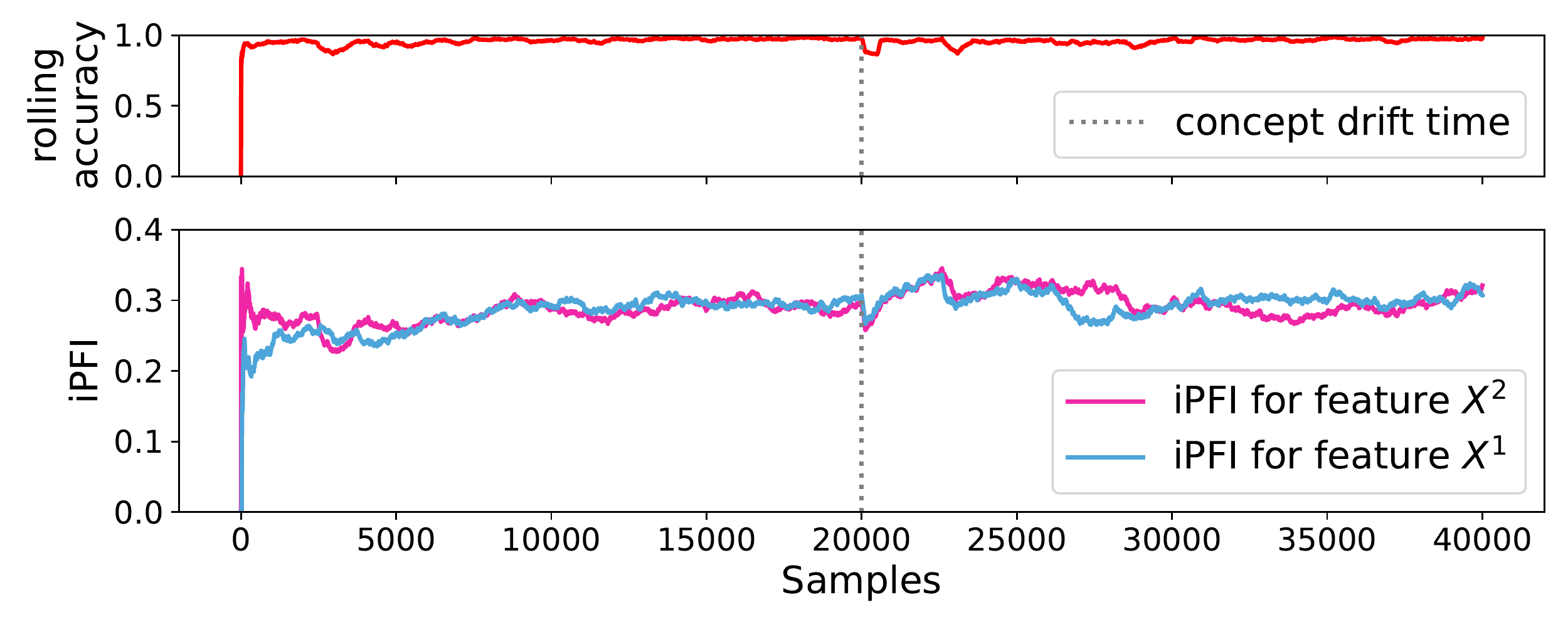}
    \caption{iPDP and iPFI for an HAT model which is fitted over time on the synthetic \emph{hyperplane} data stream. Unlike iPFI, iPDP can efficiently recover the change in feature effect direction and feature space (left to right: positive effect into negative effect).}
    \label{fig_sanity_check}
\end{figure}

Similar to the previous example, we apply iPDP in different data stream scenarios.
To illustrate the advantage of feature effects over mere FI values, we create a synthetic data stream called \emph{hyperplane}, which refers to a simple classification function described by a rotating hyperplane.
We will induce concept drift on this data stream and show that the FI values do not allow to detect the concept drift, while our iPDP estimates reveal the changes in the model.
All observations on one side of the hyperplane are considered to be of class $\textbf{1}$ and otherwise $\textbf{0}$.
Let $X^{(1)} \sim \mathcal N(\mu_1,\sigma_1^2)$, $X^{(2)} \sim \mathcal N(\mu_2,\sigma_2^2)$ and error term $\epsilon \sim \mathcal N(\mu_\epsilon,\sigma_\epsilon^2)$ be random variables.
We then define 
\begin{equation*}
    Z = \beta_1 X^{(1)} + \beta_2 X^{(2)} + \epsilon \text{ and } Y = \mathbf{1}(\frac{1}{1+\exp(-Z)} \geq \tau),
\end{equation*}
where $\beta_1,\beta_2 \in \mathbb{R}$ are scaling parameters, $0<\tau<1$ is a threshold value and $\mathbf{1}$ the indicator function, i.e. $1$ if the condition is fulfilled and $0$ otherwise.
We initialize $\tau = 0.1, \mu_1 = 100, \mu_2 = 200, \sigma_1^2 = 20, \sigma_2^2 = 40, \beta^{(1)} = 1,$ and $\beta^{(2)} = -0.5$ and induce a concept drift after $20\,000$ observations by switching to $\mu_1 = 200, \mu_2 = 100, \sigma_1^2 = 40, \sigma_2^2 = 20, \beta^{(1)} = -0.5,$ and $\beta^{(2)} = 1$.
We incrementally train a Hoeffding Adaptive Tree (HAT) \cite{Hulten.2001} on this concept drift data stream and explain it with iPDP.
The smoothing parameter is set to $\alpha = 0.001$ and 20 equidistant grid points are spread out between the minimum and maximum values of the last $2\,000$ samples (cf. Algorithm~\ref{alg_extrem_value_storage}).
Moreover, we explain this HAT by measuring its incremental permutation feature importance (iPFI) \cite{Fumagalli.2022} for this stream.
iPFI computes the well-established PFI \cite{Breiman.2001} score similar to iPDP for a non-stationary model and streaming data.
Fig.~\ref{fig_sanity_check} shows, how the drastic concept drift may be detected by drift detection mechanism based on the model's performance or its iPFI scores.
Yet, neither the change in the feature range (moving from $\mu_1 = 100$ to $\mu_1 = 200$), nor the switch in the classification function (moving from $\beta^{(1)} = 1$ to $\beta^{(1)} = -0.5$) can be derived from single point importance values.
For a further example on the same data stream illustrated in Fig.~\ref{fig_pdp_on_stream}, we refer to Appendix~\ref{sec_appendix_experiment_static}.

\subsection{Explaining in Static Environments}
\label{section_experiment_static}

\begin{figure}[t]
    \centering
    \includegraphics[width=0.6\textwidth]{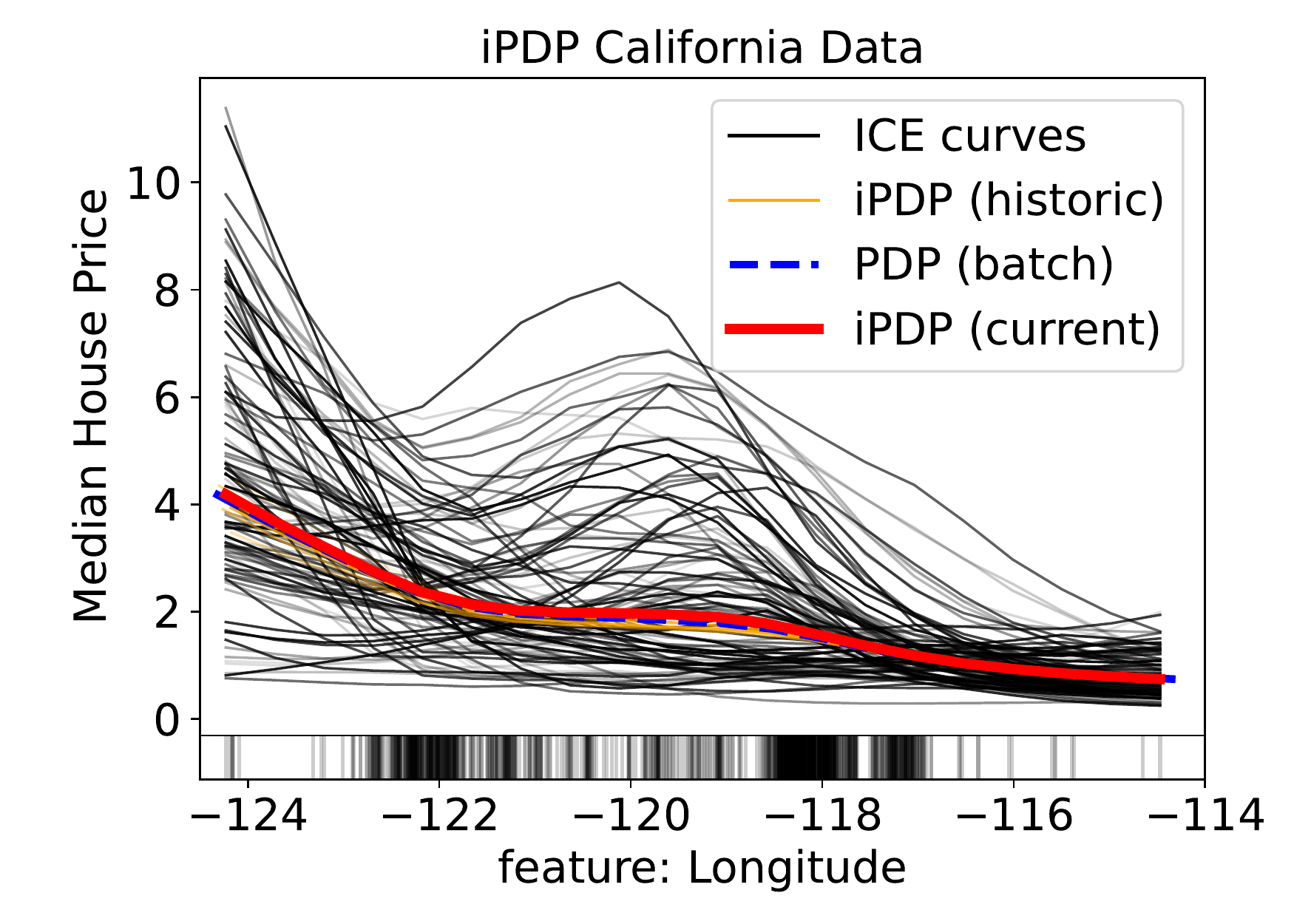}
    \caption{iPDP (red) with $\alpha = 0.01$ and batch PDP (blue, dashed) for the \emph{Longitude} feature of a NN trained on the \emph{california} housing dataset. For both iPDP and PDP, 20 equidistant grid points are selected between the minimum and maximum value of the feature's range.}
    \label{fig_static_california}
\end{figure}

To validate the theoretical results, we compare iPDP and the traditional batch PDP in static modeling scenarios.
We pre-compute different models in batch mode and then compute the traditional PDPs on the training data.
We turn the same same dataset into a data stream and compute iPDP.
For an illustration, we fit a three layer NN (50, 100, and 5 neurons) regression model on the \emph{california} housing data set.
The regressor achieves a $R^2$ of $0.806$ and a MSE of $0.077$ on a $70\%/30\%$ train-test split with standard scaled features and log-scaled (base 10) prediction targets.
The corresponding iPDP and PDP for the \emph{Longitude} feature are depicted in Fig.~\ref{fig_static_california}.
Both, iPDP and PDP, show almost identical feature effects over the same feature range.
Both methods are computed with the same amount of model evaluations.
The model has learned a, on average, positive relationship between the westwardness (more negative longitude) and the value of a property.
This trend follows similar experiments conducted on this data set \cite{Herbinger.2022}.

\section{Conclusion and Future Work}

We propose and analyze iPDP, a novel, model-agnostic XAI framework to compute feature effect plots in dynamic learning environments.
Therein, iPDP is based on the well-established PDP for static models and datasets.
We formally analyze iPDP and show that it approximates a time-dependent variant of the PDP that properly reacts to real and virtual concept drift. The time-sensitivity of iPDP is controlled by a single smoothing parameter $\alpha$, which directly corresponds to the variance and the approximation error of iPDP in a static learning environment.
In essence, iPDP transforms a model which is trained on an evolving data stream in a stream of explanations.
We have demonstrated that this explanation stream can be analyzed with traditional online learning tools to detect and investigate behavior changes in dynamic models.
We have further shown that such monitoring systems can detect changes in the model, which might otherwise be concealed in single valued importance scores.
However, computing iPDP for every feature with high fidelity (larger grid sizes) might become infeasible for some application domains.
In this case, a combination of iPDP and computationally more efficient approaches needs to be explored.


\bibliography{references}

\appendix
\section*{Organization of the Appendix}

The supplement material is organized as follows.
First, Section~\ref{appendix_proofs} contains proofs for the theoretical results described in Section~\ref{section_iPDP_theory}.
Second, Section~\ref{appendix_storage} describes the storage mechanisms introduced in Section~\ref{section_iPDP}.
Lastly, Section~\ref{sec_appendix_experiments} contains additional information about the datasets and implementation of the experiments.

\section{Proofs}\label{appendix_proofs}
\subsection{Proof of Theorem \ref{theorem_expectation}}
\begin{proof}
Using the recursion, it can be shown that $\hat f_S^{\text{PD}}(\mathbf{x}^S_{t,k},t)$ has the following explicit form
\begin{equation*}
    \hat f_S^{\text{PD}}(\mathbf{x}^S_{t,k},t) = \alpha \sum_{s=1}^t (1-\alpha)^{t-s} f_s(\tilde{\mathbf{x}}^S_{s,k},\mathbf{x}^{\bar S}_s),
\end{equation*}
where $\tilde{\mathbf{x}}^{S}_{s,k}$ is the $k$-th model evaluation point at time $s$ and $\mathbf{x}^{\bar S}_s$ is the observation at time $s$.
By the linearity of the expectation, we thus have
 \begin{align*}
    \mathbb{E}[\hat f_S^{\text{PD}}(\mathbf{x}^S_{t,k},t)] &= \alpha \sum_{s=1}^t (1-\alpha)^{t-s} \mathbb{E}_{X^{\bar S}_s}\left[f_s(\tilde{\mathbf{x}}^S_{s,k},X^{\bar S}_s)\right],
\end{align*}
which finishes the proof.
\end{proof}

\subsection{Proof of Theorem \ref{theorem_static_fixed_expectation}}
\begin{proof}
    By Theorem \ref{theorem_expectation}, the expectation of $\hat f_S^{\text{PD}}(\mathbf{x}^S_{t,k},t)$ is given as
    \begin{equation*}
         \mathbb{E}[\hat f_S^{\text{PD}}(\mathbf{x}^S_{t,k},t)] =\alpha \sum_{s=1}^t (1-\alpha)^{t-s} \mathbb{E}_{X^{\bar S}_s}\left[f_s(\tilde{\mathbf{x}}^S_{s,k},X^{\bar S}_s)\right].
    \end{equation*}
     Due to static environment assumption, we have $f_S^{\text{PD}}(\tilde{\mathbf{x}}^S_{s,k}) \equiv \mathbb{E}_{X_s^{\bar S}}[f_s(\tilde{\mathbf{x}}^S_{s,k},X_s^{\bar S})]$ for $s=1,\dots,t$.
    As the sequence of model evaluation points is constant, i.e. $\tilde{\mathbf{x}}^S_{k} \equiv \tilde{\mathbf{x}}^S_{s,k}$ for $s=1,\dots,t$, we have
        \begin{align*}
         \mathbb{E}[\hat f_S^{\text{PD}}(\mathbf{x}^S_{t,k},t)] &=\alpha \sum_{s=1}^t (1-\alpha)^{t-s} \mathbb{E}_{X^{\bar S}_s}\left[f(\tilde{\mathbf{x}}^S_{k},X^{\bar S}_s)\right] = \alpha \sum_{s=1}^t (1-\alpha)^{t-s} f_S^{\text{PD}}(\tilde{\mathbf{x}}^S_{k}) 
         \\
         &= (1-(1-\alpha)^t) f_S^{\text{PD}}(\tilde{\mathbf{x}}^S_{k}),
    \end{align*}
    where we used the properties of a finite geometric series.
\end{proof}

\subsection{Proof of Theorem \ref{theorem_static_expectation}}
\begin{proof}
    By Theorem \ref{theorem_expectation}, the expectation of $\hat f_S^{\text{PD}}(\mathbf{x}^S_{t,k},t)$ is given as
    \begin{equation*}
         \mathbb{E}[\hat f_S^{\text{PD}}(\mathbf{x}^S_{t,k},t)] =\alpha \sum_{s=1}^t (1-\alpha)^{t-s} \mathbb{E}_{X^{\bar S}_s}\left[f_s(\tilde{\mathbf{x}}^S_{s,k},X^{\bar S}_s)\right].
    \end{equation*}
    Due to static environment assumption, we have $f_S^{\text{PD}}(\tilde{\mathbf{x}}^S_{s,k}) \equiv \mathbb{E}_{X_s^{\bar S}}[f(\tilde{\mathbf{x}}^S_{s,k},X_s^{\bar S})]$ for $s=1,\dots,t$ and thus by local linearity of $f$ and linearity of expectation
    \begin{align*}
        \mathbb{E}[\hat f_S^{\text{PD}}(\tilde{\mathbf{x}}^S_{t,k},t)]&=\alpha \sum_{s=1}^{t} (1-\alpha)^{t-s}f_S^{\text{PD}}(\tilde{\mathbf{x}}^S_{s,k})
        \\
        &=\mathbb{E}_{X^{\bar S}}[\alpha \sum_{s=1}^{t} (1-\alpha)^{t-s} f(\tilde{\mathbf{x}}^S_{s,k},X^{\bar S}])
        &\text{(linearity of expectation)}
        \\
        &=\mathbb{E}_{X^{\bar S}}[f(\alpha \sum_{s=1}^{t} (1-\alpha)^{t-s}\tilde{\mathbf{x}}^S_{s,k},X^{\bar S})]) &\text{(local linearity)}
        \\
        &=\mathbb{E}_{X^{\bar S}}[f(\mathbf{x}^S_{t,k},X^{\bar S})])
        \\
        &= f_S^{\text{PD}}(\mathbf{x}^S_{t,k}),
    \end{align*}    
    where we have used the explicit form of the grid points
    \begin{equation*}
        \mathbf{x}^S_{t,k} = \alpha \sum_{s=1}^t (1-\alpha)^{t-s}\tilde{\mathbf{x}}^S_{s,k},
    \end{equation*}
    obtained from the exponential moving average in (\ref{equation_grid_points_recursion}).
    Multiplying both sides of the equation with the debiasing factor $(1-(1-\alpha)^t)^{-1}$ and using linearity of the expectation and local linearity of $f$ yields the debiased variant.
\end{proof}

\subsection{Proof of Theorem \ref{theorem_static_variance}}
\begin{proof}
    As all observations are drawn independently, the variance can be computed as
    \begin{align*}
    \mathbb{V}[\hat f_S^{\text{PD}}(\mathbf{x}^S_{t,k},t)] = \alpha^2\sum_{s=1}^t (1-\alpha)^{2(t-s)} \mathbb{V}[f_s(\tilde{\mathbf{x}}^S_{s,k},X^{\bar S}_s)] \leq \alpha^2\sum_{s=1}^t (1-\alpha)^{2(t-s)} \sigma^2_k,
    \end{align*}
    where $\sigma^2_k$ is the maximum variance in the range of temporary model evaluation points with $f \equiv f_s$ and $X \equiv X_s$.
    By the properties of the geometric series, we can bound
    \begin{align*}
        \mathbb{V}[\hat f_S^{\text{PD}}(\mathbf{x}^S_{t,k},t)] &= \sigma^2_k  \alpha^2\sum_{s=1}^t (1-\alpha)^{2(t-s)} \leq \sigma^2_k  \alpha^2\sum_{s=0}^\infty (1-\alpha)^{2s} = \sigma_k^2 \frac{\alpha}{2-\alpha} 
        \\
        &= \mathcal O (\alpha),
    \end{align*}
    which finishes the first part of the proof.
    By the local linearity assumption, we have by Theorem \ref{theorem_static_expectation} that $\mathbb{E}[\hat f_S^{\text{PD}}(\mathbf{x}^S_{t,k},t)]=f_S^{\text{PD}}(\mathbf{x}^S_{t,k})$.
    Hence, by Chebyshev's inequality for $\epsilon>0$
    \begin{align*}
    \mathbb{P}(\vert \hat f_S^{\text{PD}}(\mathbf{x}^S_{t,k},t)-f_S^{\text{PD}}(\bar{\mathbf{x}}^S_{t,k})\vert > \epsilon) &= \mathbb{P}(\vert \hat f_S^{\text{PD}}(\mathbf{x}^S_{t,k},t)-\mathbb{E}[\hat f_S^{\text{PD}}(\mathbf{x}^S_{t,k},t)]\vert > \epsilon)
    \\
    &\leq \frac{\mathbb{V}[\hat f_S^{\text{PD}}(\mathbf{x}^S_{t,k},t)]}{\epsilon^2} = \mathcal O (\alpha).
    \end{align*}
    
\end{proof}

\section{Efficient Storage Mechanisms of Feature Values}
\label{appendix_storage}

This section contains further details on the two storage mechanisms introduced in Section~\ref{section_iPDP}.
Section~\ref{app_extreme_values_storage} contains the storage mechanism for storing extreme values of the feature range of a window in a data stream.
Section~\ref{app_frequency_distribution_storage} contains the storage mechanism for storing the frequency distribution of a feature of a window in a data stream.

\subsection{Storing Extreme Values of the Feature Range}
\label{app_extreme_values_storage}

\begin{figure}[t]
    \centering
    \includegraphics[width=\textwidth]{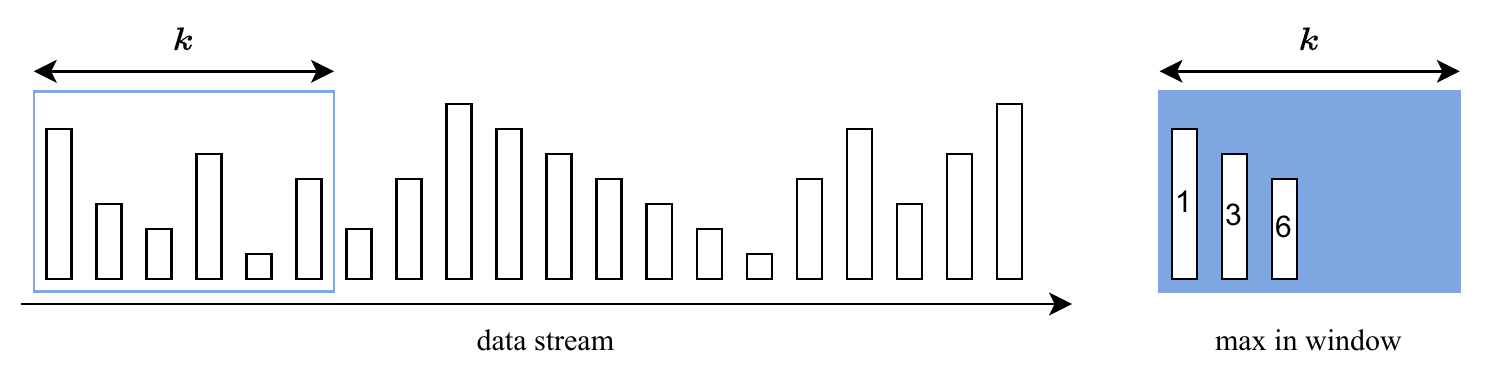}
    \textbf{(a) expected case with iid data}\\[1em]
    \includegraphics[width=\textwidth]{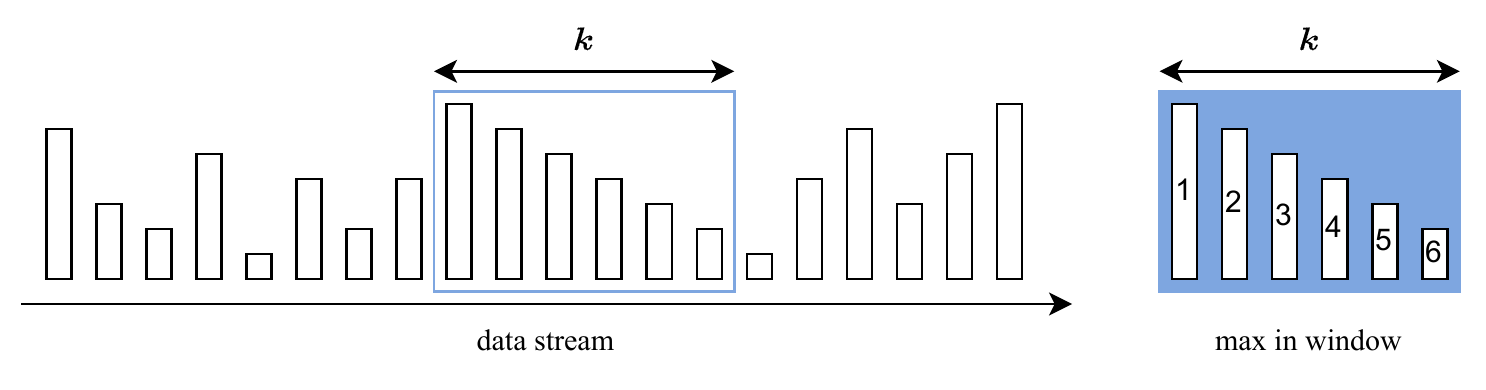}
    \textbf{(b) worst case scenario}\\[1em]
    \caption{Illustration of the extreme value storage mechanism where the current max value is to be stored in a data stream. The sorted window of a maximum length $k$ keeps track of the current max values and how long the max values remain in storage (numbers in the bars). The first bar from the left in the storage reservoir is the current max value. With iid data (a), only a few data points need to be stored to reliably maintain the current max value. However, if values are decreasing strictly monotonic (b) then the reservoir needs to be filled completely. 
    }
    \label{fig_illustration_min_max_storage_mechanism}
\end{figure}

The rolling minimum $\mathbf{x}^S_{t,\min}$ and maximum $\mathbf{x}^S_{t,\max}$ at every time step $t$ over a window of length $w$ are maintained using an efficient storage mechanism, illustrated in Fig.~\ref{fig_illustration_min_max_storage_mechanism}.
Our efficient implementation of the rolling maximum is based on the observation that previous data points with lower values than a more recent observation can be discarded.
The reverse argument holds for the rolling minimum.
While this algorithm has worst-case storage requirements of the whole window length, it is in practice substantially more memory efficient.
The implementation for storing maximum values is outlined in Algorithm~\ref{alg_extrem_value_storage} (storing minimum values can be achieved by multiplying $x_t$ with $-1$).
The procedure depicts the general strategy of the storage system. 
For run-time efficiency, specific data structures are required, which depend on the implementation environment. A \emph{Python} implementation can be found in the technical supplement. The storage strategy is further illustrated in Fig.~\ref{fig_illustration_min_max_storage_mechanism}.

\begin{algorithm}[t]
\caption{Incremental Extreme Value Storage}
\label{alg_extrem_value_storage}
\begin{algorithmic}[1]
\REQUIRE stream of feature values $\left\{x_{t}\right\}_{t=1}^{\infty}$, sorted storage reservoir $R$ of size $k$.
\STATE Initialize $R \gets \emptyset$
\FORALL{$x_{t} \in$ stream}
    \STATE $x_{\text{max}} \gets \text{max}(R_{t-1})$ \COMMENT{get current max value}
    \STATE $t_{\text{inserted}} \gets \textsc{GetAgeInReservoir}(x_{\text{max}})$ \COMMENT{get age of current max value}
    \IF{$k - t_{\text{inserted}} \leq 0$}
        \STATE $R \gets R \setminus \{x_{\text{max}}\}$ \COMMENT{remove outdated max value based on age}
    \ENDIF
    \STATE $\left\{x_{i} \vert x_i \leq x_t \right\} \gets \textsc{Smaller}(R, x_{t})$ \COMMENT{get all values smaller or equal to $x_t$}
    \STATE $R \gets R \setminus \left\{x_{i} \vert x_i \leq x_t \right\}$ \COMMENT{remove all values smaller than $x_t$ from the reservoir}
    \STATE $R \gets R \cup \{x_{t}\}$ \COMMENT{add value to the end of the reservoir}
\ENDFOR
\end{algorithmic}
\end{algorithm}

\subsection{Storing Feature Frequency Distributions}
\label{app_frequency_distribution_storage}

Efficiently storing the frequency distribution of a window with length $w$ in a data stream is not possible with traditional streaming histograms \cite{BenHaim.2010} as old observations outside the window's range need to be forgotten. 
Yet, the \emph{geometric reservoir storage} mechanism \cite{Fumagalli.2022} introduced to store the recent data distribution can be used and further optimized to efficiently model the frequency distribution of a window.
Originally, a fixed-length reservoir with size $L$ is updated with each new observation.
A new observation always replaces an old observation from the reservoir, where the old data point is chosen uniformly at random.
This leads to the reservoir containing data points, where the probability of belonging to the reservoir follows a geometric distribution in terms of recency.
However, to hold information about large window sizes, the size of the reservoir may be limiting.

The space requirement can be further optimized by changing the entrance probability of a new observation being added to the reservoir and the selection criterion for choosing the observation to be dropped from the reservoir.
Including a new observation with probability $0<p_{\text{inc}} <1$ to the reservoir, increases the slows the replacement of previous samples in the reservoir.
Moreover, if the replacement samples are not selected uniformly but based on their time spent in the reservoir or as a \emph{Last-In-First-Out} (LIFO) queue, older samples can be reliably removed. 
With this, a small reservoir of size $L$ can be used to compute the frequency distribution of the last $w = \nicefrac{L}{P(E)}$ window of samples.
The procedure is illustrated in 
Algorithm~\ref{alg_geom_reservoir_storage_efficient}.

\begin{algorithm}
\caption{Incremental Frequency Storage}
\label{alg_geom_reservoir_storage_efficient}
\begin{algorithmic}[1]
\REQUIRE stream of feature values $\left\{x_{t}\right\}_{t=1}^{\infty}$, storage reservoir $R$ of size $L$, and an entrance probability $0 < P(E) \leq 1$.
\STATE Initialize $R \gets \emptyset$
\FORALL{$x_{t} \in$ stream}
    \IF{$\vert R \vert < L$} 
        \STATE $R \gets R \cup \{x_t\}$ \COMMENT{add $x_r$ to the reservoir}
    \ENDIF
    \STATE $V \sim \text{Unif}(0,1)$
    \IF{$V \leq P(E)$} 
        \STATE $x_r \gets \textsc{Select}(R)$ \COMMENT{select replacement value $x_r$ from $R$ deterministicly (e.g., always the oldest value) or sample it based on the age}
        \STATE $R \gets R \setminus \{x_r\}$ \COMMENT{remove $x_r$ from the reservoir}
        \STATE $R \gets R \cup \{x_t\}$ \COMMENT{add $x_r$ to the reservoir}
    \ENDIF
\ENDFOR
\end{algorithmic}
\end{algorithm}


\section{Experiments}
\label{sec_appendix_experiments}

\subsection{Data Set Description}
\label{sec_appendix_data_set_description}

\paragraph{california}
Regression dataset containing $20\,640$ samples of 8 numerical features with 1990 census information from the US state of California. 
The dataset is available at \url{https://www.dcc.fc.up.pt/~ltorgo/Regression/cal_housing.html}

\paragraph{agrawal}
Synthetic data stream generator to create binary classification problems to decide whether an individual will be granted a loan based on nine features, six numerical and three nominal.
There are ten different decision functions available.
\emph{agrawal} is a publicly available dataset \cite{Agrawal.1993}.

\paragraph{elec2}
Binary classification dataset that classifies, if the electricity price will go up or down.
The data was collected for $45312$ time stamps from the Australian New South Wales Electricity Market and is based on eight features, six numerical and two nominal.
The data stream contains a well-documented concept drift in its \emph{vicprice} feature in that the feature has no values apart from zero in all observations up to $\approx20\,000$ samples.
After that the \emph{vicprice} feature starts having values different from zero.
\emph{elec2} is a publicly available dataset \cite{elec2.1999}.

\paragraph{hyperplane}
Binary classification data stream that classifies, weather data points are on one side or the other of a hyperplane. The data stream is part of the technical supplement and will be made publicly available.

\subsection{Further Experiments in Dynamic Learning Environments}
\label{sec_appendix_experiment_static}

\begin{figure}
    \centering
    \begin{minipage}[c]{0.32\textwidth}
        \includegraphics[width=\textwidth]{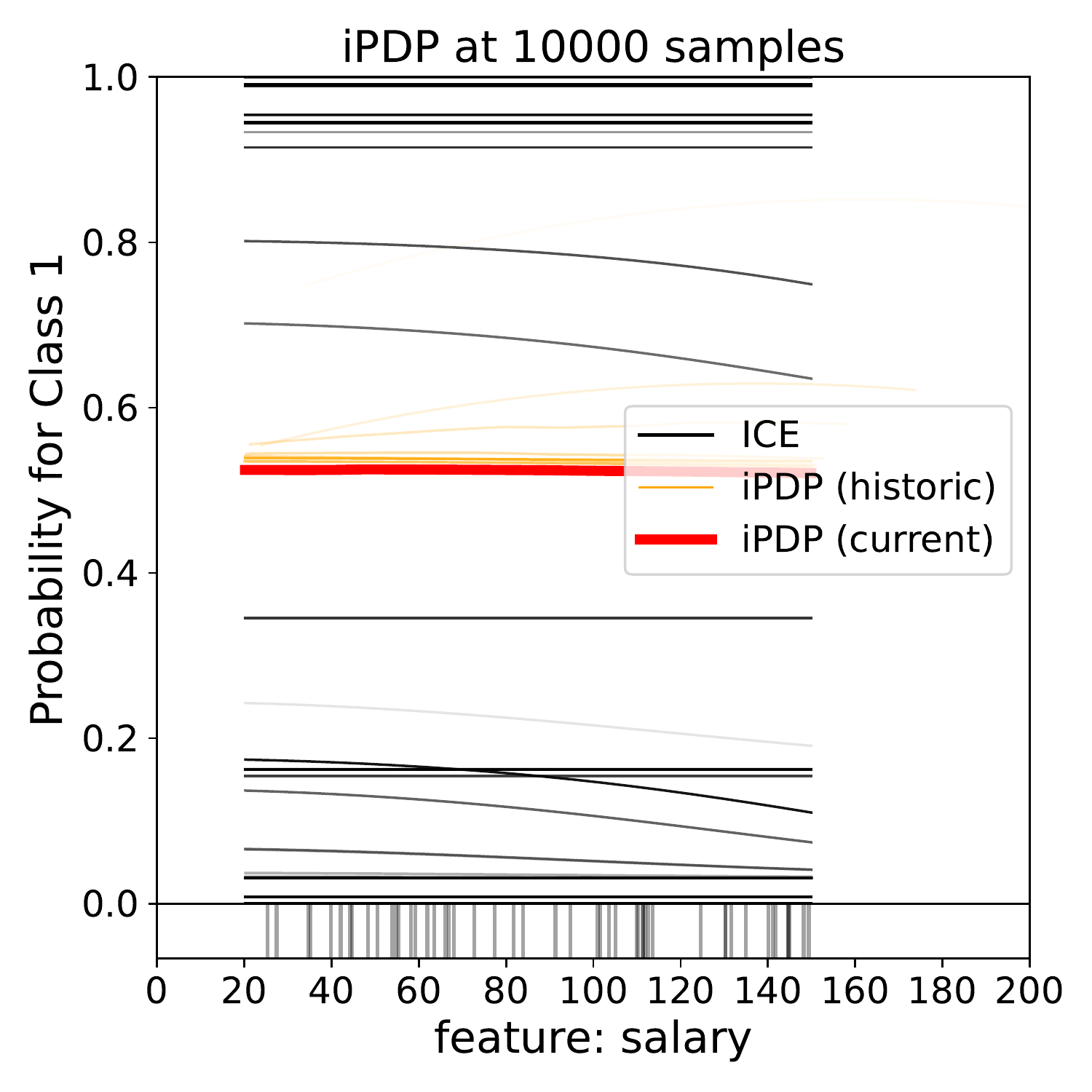}
    \end{minipage}
    \hfill
    \begin{minipage}[c]{0.32\textwidth}
        \includegraphics[width=\textwidth]{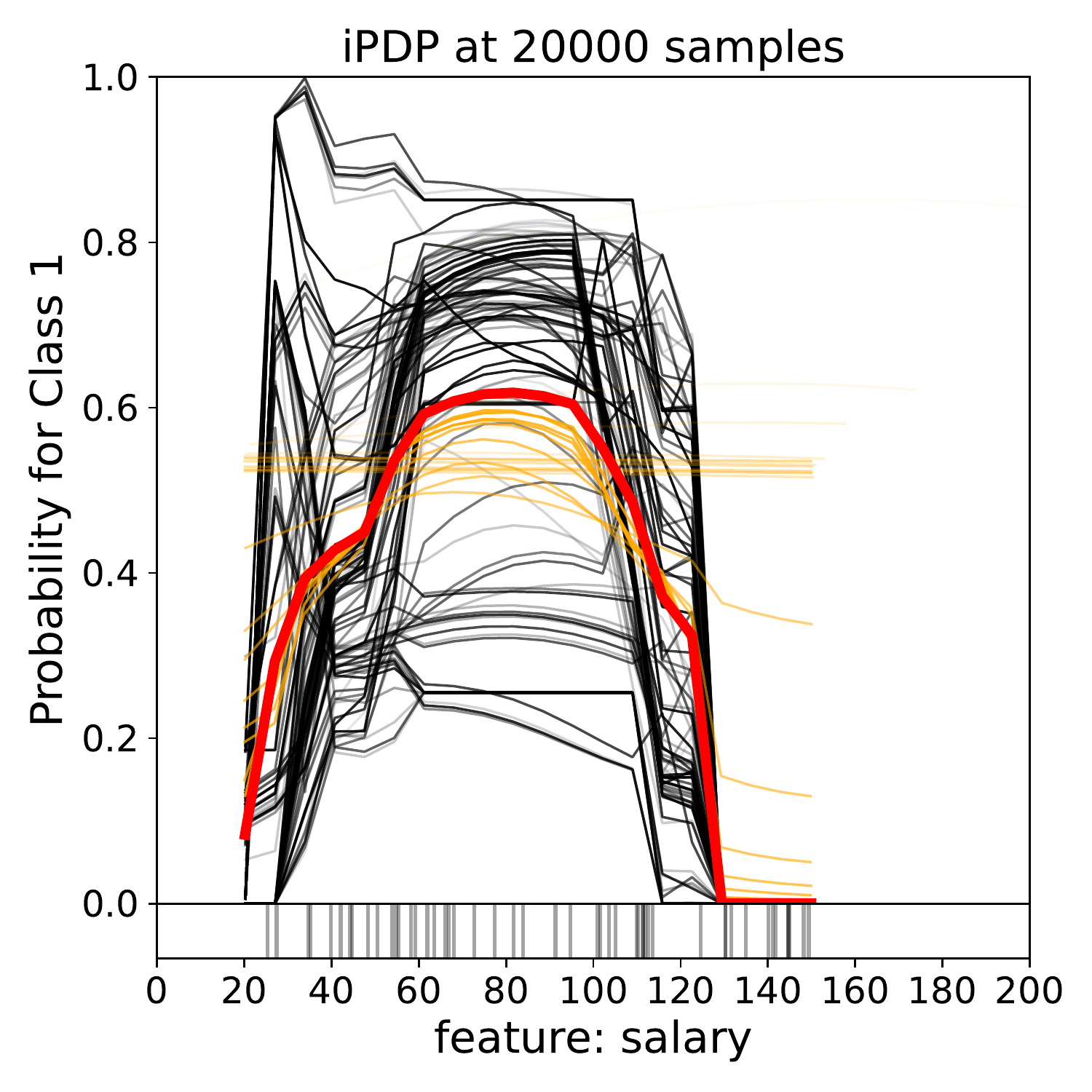}
    \end{minipage}
    \hfill
    \begin{minipage}[c]{0.32\textwidth}
        \includegraphics[width=\textwidth]{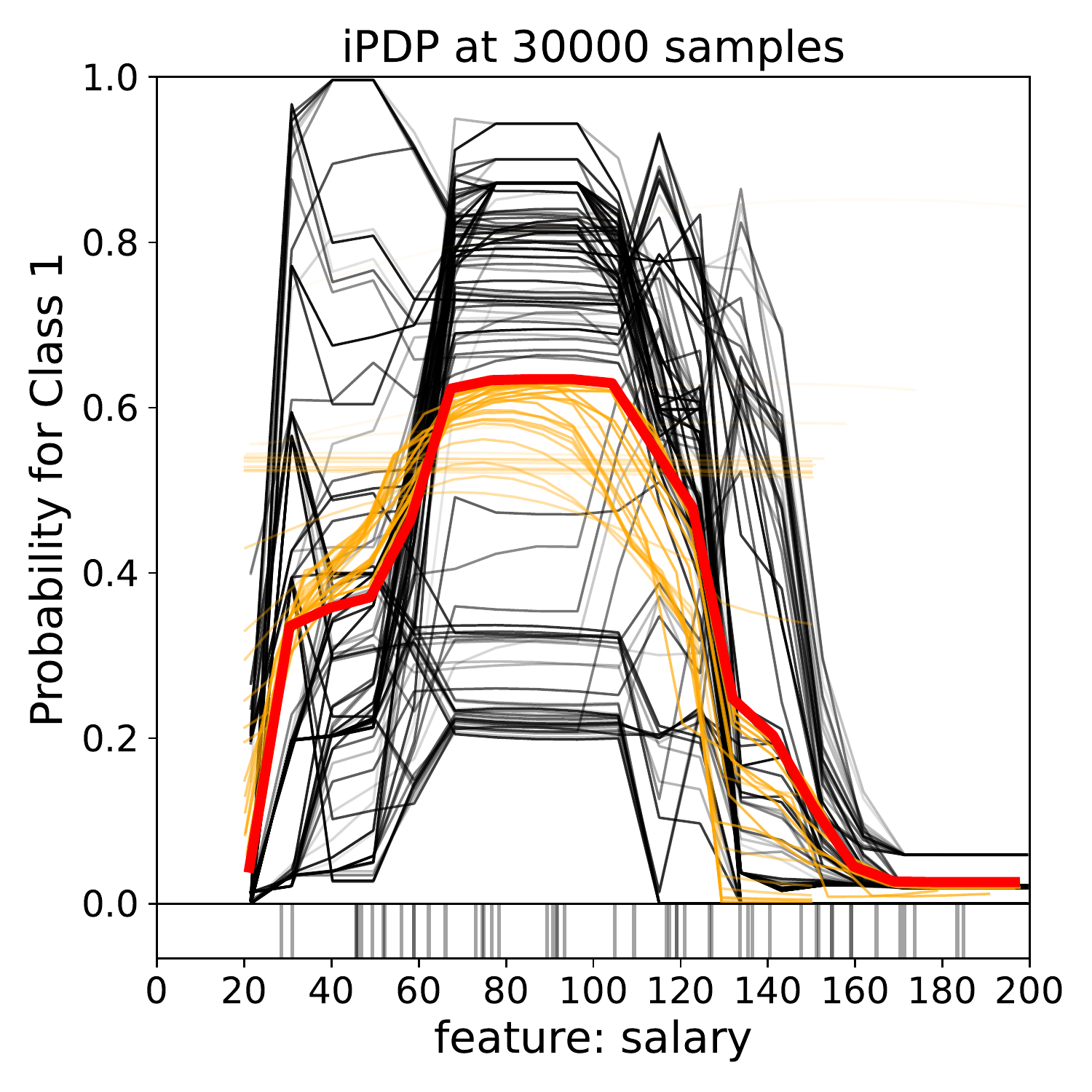}
    \end{minipage}
    \includegraphics[width=\textwidth]{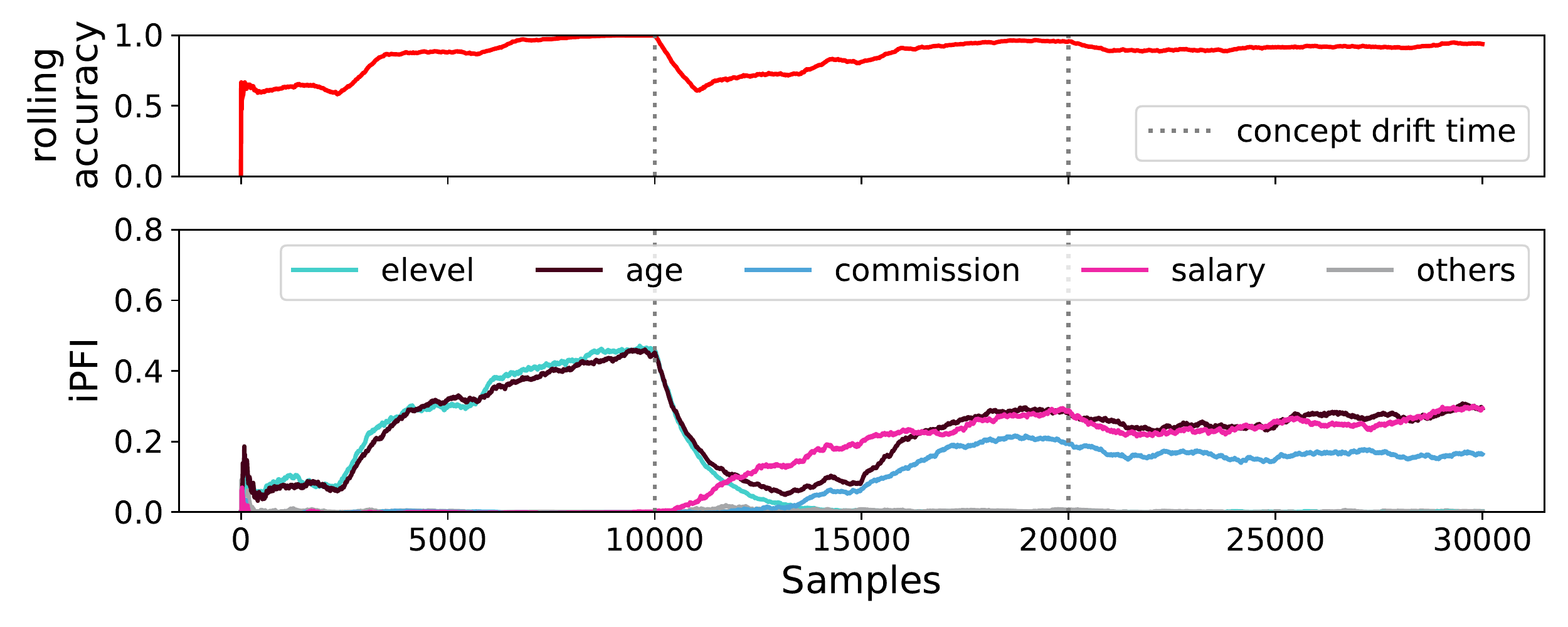}
    \caption{iPDP and iPFI for an ARF model fitted on the \emph{agrawal} \protect{\cite{Agrawal.1993}} concept drift stream illustrated in Fig.~\protect{\ref{fig_pdp_on_stream}}. The smoothing factor for both iPDP and iPFI is set to $\alpha = 0.001$.}
    \label{fig_iPDP_agrawal_concept_drift}
\end{figure}

Similar to the experiment in Section~\ref{section_experiment_dynamic_environments}, in Fig.~\ref{fig_iPDP_agrawal_concept_drift} we compute iPDP and iPFI for the same \emph{agrawal} \cite{Agrawal.1993} concept drift stream used for computing the batch PDP in Fig.~\ref{fig_pdp_on_stream}.
We incrementally fit an ARF with 10 base learners on the stream. 
The iPDPs in Fig.~\ref{fig_sanity_check} show the same feature effects as the batch PDP in Fig.~\ref{fig_pdp_on_stream}.

\end{document}